\def\year{2022}\relax
\documentclass[letterpaper]{article} 
\usepackage{aaai22}  
\usepackage{times}  
\usepackage{helvet}  
\usepackage{courier}  
\usepackage[hyphens]{url}  
\usepackage{graphicx} 
\def\UrlFont{\rm}  
\usepackage{natbib}  
\usepackage{caption} 
\DeclareCaptionStyle{ruled}{labelfont=normalfont,labelsep=colon,strut=off} 
\usepackage{algorithm}
\usepackage{algorithmic}
\usepackage[switch]{lineno}
\usepackage{amsmath}
\usepackage{amsfonts}
\usepackage{multirow}

\usepackage{newfloat}
\usepackage{listings}
\floatstyle{ruled}
\newfloat{listing}{tb}{lst}{}
\floatname{listing}{Listing}
\title{

GALAXY: A Generative Pre-trained Model for Task-Oriented Dialog with \\  Semi-Supervised Learning and Explicit Policy Injection
}

\author{Wanwei He\textsuperscript{1,2,3}\thanks{Equal Contribution.}\thanks{Work done while the author was interning at Alibaba.}, Yinpei Dai\textsuperscript{3}\footnotemark[1], Yinhe Zheng\textsuperscript{3}, Yuchuan Wu\textsuperscript{3}, Zheng Cao\textsuperscript{3}, Dermot Liu\textsuperscript{3} \\Peng Jiang\textsuperscript{3}, Min Yang\textsuperscript{1}\thanks{Corresponding authors.}, Fei Huang\textsuperscript{3}, Luo Si\textsuperscript{3}, Jian Sun\textsuperscript{3}, Yongbin Li\textsuperscript{3}\footnotemark[3]}
\affiliations{
\textsuperscript{1}Shenzhen Institute of Advanced Technology, Chinese Academy of Sciences, China\\
\textsuperscript{2}University of Chinese Academy of Sciences, China\\
\textsuperscript{3}Alibaba Group\\
\{ww.he, min.yang\}@siat.ac.cn, 
\{yinpei.dyp, shuide.lyb, f.huang, luo.si, jian.sun\}@alibaba-inc.com}

\begin{document}
\maketitle
\begin{abstract}
\begin{quote}
Pre-trained models have proved to be powerful in enhancing task-oriented dialog systems. 
However, current pre-training methods mainly focus on enhancing dialog understanding and generation tasks while neglecting the exploitation of dialog policy.
In this paper, we propose GALAXY, a novel pre-trained dialog model that explicitly learns dialog policy from limited labeled dialogs and large-scale unlabeled dialog corpora via semi-supervised learning.
Specifically, we introduce a dialog act prediction task for policy optimization 
during pre-training and employ a \textit{consistency regularization} term to refine the learned representation with the help of unlabeled dialogs.
We also implement a gating mechanism to weigh suitable unlabeled dialog samples.
Empirical results show that GALAXY substantially improves the performance of task-oriented dialog systems, and achieves new state-of-the-art results on benchmark datasets: In-Car, MultiWOZ2.0  and  MultiWOZ2.1,  improving their end-to-end combined scores by 2.5, 5.3 and 5.5 points, respectively.
We also show that GALAXY has a stronger few-shot ability than existing models under various low-resource settings. 
For reproducibility, we release the code and data at \UrlFont{https://github.com/siat-nlp/GALAXY}.
\end{quote}
\end{abstract}

\section{Introduction}
\begin{figure}
    \centering
    \includegraphics[width=0.4\textwidth]{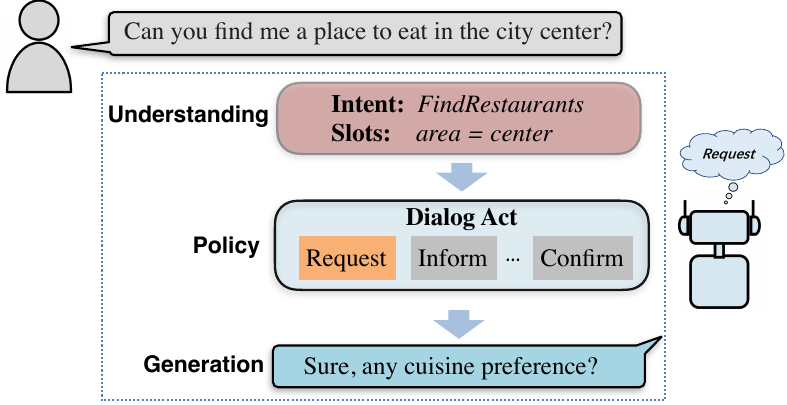}
    \caption{Given the input user utterance, a task-oriented dialog system needs to perform understanding, policy planning, and generation successively to complete  the reply.}
    \label{fig:example}
\end{figure}

Task-oriented dialog (TOD) systems aim to help users accomplish certain tasks through conversations. Fundamental abilities of a TOD system include: (1) \textit{Dialog understanding}: extracting structured semantics from user utterances; (2) \textit{Policy planning}: determining a \underline{D}ialog \underline{A}ct (DA) that leads to successful task completion; and (3) \textit{Dialog generation}: producing appropriate responses (Figure \ref{fig:example}).
With the recent progress of \underline{P}re-trained \underline{L}anguage \underline{M}odels (PLMs), remarkable performances improvements are achieved by casting TODs as generative language modeling tasks \cite{peng2020soloist, lin2020mintl}, which benefit from the rich linguistic knowledge embedded in PLMs.


However, as reported in previous studies \cite{zhang2019dialogpt, kulhanek2021augpt}, there are intrinsic differences between the distribution of human conversations and plain texts. Directly fine-tuning plain-text-trained PLMs on downstream dialog tasks hinders the model from effectively capturing conversational linguistic knowledge and thus leads to sub-optimal performances \cite{mehri2019pretraining, zenginvestigation, wu2020probing}. Current attempts to tackle this issue try to build \underline{P}re-trained \underline{C}onversation \underline{M}odels (PCMs) by directly optimizing vanilla language model objectives on dialog corpora \cite{mehri2020dialoglue,zhang2019dialogpt, henderson2019convert}, which shows improved results on both dialog understanding \cite{wu2020tod} and generation \cite{peng2020few}.

Despite these reported advances, few approaches are proposed to further enrich the pre-training process of PCMs with the  knowledge of dialog policy.
Specifically, existing methods either ignore explicit policy modeling or use latent variables without considering external dialog policy information \cite{bao2020plato}, which hinders the possibility of learning controllable policy during pre-training.
The optimization of dialog policy is usually formulated as a DA prediction task, which is crucial in TOD systems \cite{su2017sample,liu2018dialogue}.
Therefore, we hypothesize that explicitly incorporating the DA annotations into the  pre-training process can also facilitate learning better representations for policy optimization to improve the overall end-to-end performance. 
A naive way to utilize these labels is to design a multi-task learning process \cite{sun2020ernie} that directly combines vanilla unsupervised pre-training losses such as MLM \cite{devlin2018bert} with a supervised DA classification loss.
However, this approach has several drawbacks when generalizing to large-scale pre-training paradigms: 
(1) The DA annotation schema is inconsistent among existing corpora, making it challenging to collect large-scale DA annotations;
(2) A vast majority of available dialogs do not have DA labels. A naive joint training process without careful regularization would lead to highly over-fitting on those labeled samples, resulting in low performance;
(3) All supervision signals from unlabeled data are self-supervised without any explicit inference over the DA space, so the linguistic knowledge PCMs can extract is only the general type, and the knowledge of dialog policy can not be effectively explored. 





In this study, we propose a novel generative pre-trained model called GALAXY, aiming to inject the knowledge of dialog policy explicitly into pre-training at low cost while maintaining its strong ability on dialog understanding and generation.  
To begin with,  we build a unified DA taxonomy for TOD and examine eight existing datasets to  develop a new labeled dataset named \textit{UniDA} with a total of 975K utterances. We also collect and process a large-scale unlabeled dialog corpus called \textit{UnDial} with 35M utterances, whose scenarios ranging from online forums to customer services.
Then, we propose a  semi-supervised pre-training paradigm that applies \textit{consistency regularization} \cite{verma2019interpolation} on all data. It minimizes the bi-directional KL-divergence between model predictions made on dropout-perturbed samples, which facilitates better representation learning from unlabeled dialog corpora. 
Since a large proportion of UnDial is from the Internet and not well-suited to our DA taxonomy, we add a learnable control gate on the KL loss of unlabeled data, so that only good samples are allowed for the  consistent regularization, other samples are restricted back to normal self-supervised objectives.
Experiments show that GALAXY substantially improves TOD systems and achieves new state-of-the-art results on In-Car, MultiWOZ2.0, and MultiWOZ2.1, pushing the end-to-end combined score to 107.45, 110.35, and 110.76, respectively. We also observe that GALAXY has a strong few-shot ability under various low-resource settings.

%

In summary, our main contributions are three-fold:
\begin{itemize}
    \item To the best of our knowledge,
this is the first study to use semi-supervised pre-training to model explicit dialog policy for PCMs.
     \item Experiments show our model has learned the knowledge of dialog policy, and achieves new state-of-the-art performance on several TOD benchmarks;
     \item We collect a new labeled 
dataset \textit{UniDA} as well as a large-scale unlabeled dialog corpus \textit{UnDial}, hoping that can help bring forward the research in this area.
\end{itemize}

\section{Related Work}

\paragraph{Pre-trained Language Models (PLMs)} are trained on large-scale textual corpora with Transformer \cite{devlin2018bert, gpt2}, which significantly improve dialog systems performance.
\citet{budzianowski2019hello} is the first work to validate the possibility of fine-tuning the information of all sub-tasks in a single paragraph of text on GPT-2.
SimpleTOD \cite{hosseini2020simple}  and SOLOIST \cite{peng2020soloist} further generalize this idea to an end-to-end setting where the semantic labels are generated instead of using ground truth values and also consider database results in the training process.
\citet{yang2020ubar} leverage the entire dialog session as the input sequence and demonstrate superior performance using self-generated responses during evaluation.

\paragraph{Pre-trained Conversation Models (PCMs)} 
are variants of PLMs  particularly adapted for conversational modeling. 
The main adaptation methods can be roughly divided into three types. The first is training PLMs on dialog corpora instead of plain texts with vanilla language model objectives. 
Recent work, such as DialoGPT \cite{zhang2019dialogpt}, Meena \cite{adiwardana2020towards} and Blender \cite{roller2020recipes} are trained on billions of open-domain dialogs, demonstrating powerful dialog generation performances. 
TOD-BERT \cite{wu2020tod} shows a  great few-shot ability in various understanding tasks via pre-training BERT on extensive task-oriented dialog data.
The second line is to design new dialog-oriented pre-training objectives \cite{bao2020plato, he2020amalgamating, he2021multi, xu2021dialogue, su2021multitask, dai-etal-2021-preview}.
\citet{bao2020plato} use discrete latent variables to tackle the one-to-many mapping problem in open-domain dialog generation.
\citet{xu2021dialogue} propose to simulate the conversation features only using plain texts. 
The third is to integrate dialog annotations into the pre-training stage. \citet{yu2020score} use labels of dialog understanding as supervision to pre-train BERT.
\citet{peng2020few} use labeled conditional generation data to enhance dialog generation performance. Different from them, we are the first to utilize labels of dialog policy to improve PCMs.


\paragraph{Semi-supervised Learning (SSL)} learns from both unlabeled and labeled data. 
Approaches differ on what information to acquire from the structure of the unlabeled samples. Many initial results  were based on generative models, such as variational autoencoders \cite{kingma2019introduction}  and generative adversarial networks \cite{goodfellow2014generative}. Pseudo-Labeling \cite{lee2013pseudo} is another widely used method, where unlabeled data is used as further training data after predicted by a model trained on labeled data. One line of recent research shows promising results by jointly training labeled data with supervised learning and unlabeled data with self-supervised learning \cite{sun2020ernie}. This lies in the paradigm of multi-task learning, where lower layers are often shared across all tasks while the top layers are task-specific. Consistency regularization \cite{verma2019interpolation} is also a prominent method in SSL, which improves classification performance by minimizing the discrepancy between predictions made on perturbed unlabeled data points. 
Recently, SimCSE \cite{gao2021simcse} leverages dropout as the perturbed method and uses a contrastive objective as the regularization loss to learn sentence representations.
Inspired by SimCSE, we adopt the same dropout method for perturbation, and use the bidirectional KL-divergence as in  \citet{liang2021r} as our regularization loss, hoping to learn better representations that encodes the knowledge of dialog policy for downstream tasks. There are also some works \cite{jin2018explicit, zhang-etal-2020-probabilistic, liu2021variational} focusing on using latent variable models to alleviate the reliance on dialog labels via semi-supervised learning, but our work mainly targets the semi-supervised dialog pre-training.

\section{Pre-training Dialog Datasets}

In this section, we describe the new dialog datasets used for pre-training, including a  labeled dialog dataset (\textit{UniDA}) and a large-scale unlabeled dialog corpus (\textit{UnDial}).

\subsection{Labeled Dataset: \textit{UniDA}}
Dialog policy\footnote{In some datasets, the dialog act is defined as a combination of an act  and its semantic contents. To unify different datasets, we neglect the contents and only use dialog acts as the annotations. We also focus on the text-in-text-out TOD systems in this paper, and leave the spoken DA in the future research.} 
is tasked to predict dialog acts (DAs) given dialog context. Although DAs are general tags to describe speakers'  communicative behaviors \cite{bunt2009dit},  current DA annotations in task-oriented dialog are still limited and lack of unified taxonomy because each dataset is small and scattered. Recently, \citet{paul2019towards} propose a universal task-oriented DA schema, but their dataset is still insufficient for pre-training purposes and the schema lacks some important features such as \textit{not\_sure} and \textit{dont\_understand}. To this end, we follow ISO \cite{bunt2010towards} and propose a more comprehensive unified DA taxonomy for task-oriented dialog, which consists of 20 frequently-used DAs. A complete description of the taxonomy is in Appendix A.1. Base on that, we align the annotations of eight existing benchmarks: MultiWOZ
\cite{budzianowski2018multiwoz}, Frames \cite{asri2017frames}, MSRe2e \cite{li2018microsoft}, SGD \cite{rastogi2020towards}, DSTC2 \cite{henderson2014second}, SimJoint \cite{shah2018bootstrapping}, STAR
\cite{mosig2020star} and DailyDialog \cite{li2017dailydialog}. We add DailyDialog, an open-domain dialog dataset, to accommodate our dialog policy for more general types. Finally, a new dataset \textit{UniDA} is obtained. 
Table \ref{tab:UniDA} shows more detailed statistics.

\begin{figure*}[htp]
    \centering
    \includegraphics[width=0.80\textwidth]{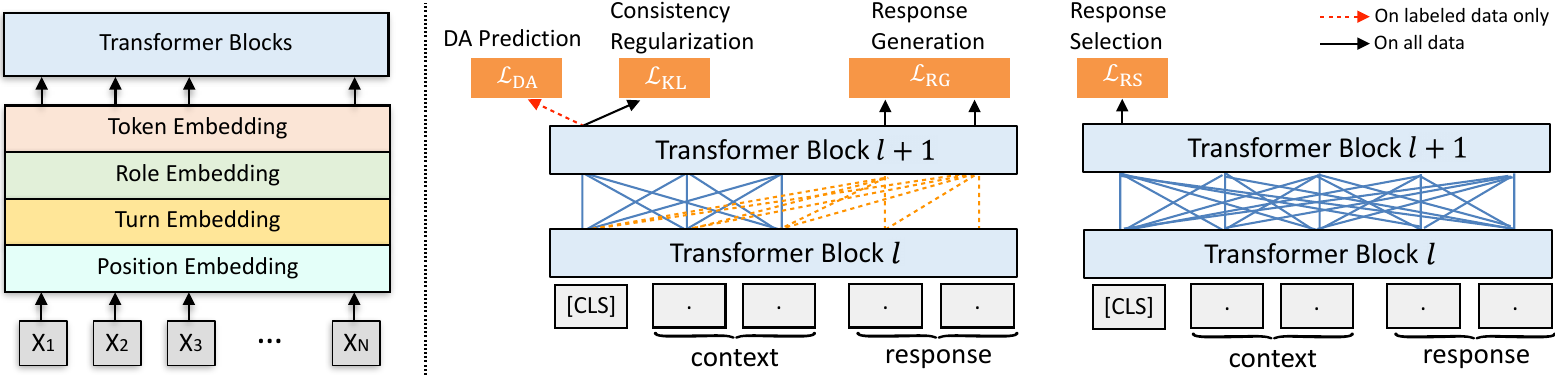}
    \caption{Architecture of our pre-trained dialog model. The left part illustrate the input representations, which contain embeddings of tokens, roles, turns, and positions. The right part shows the pre-trained objectives.
    Blue lines denote the bi-directional attention. Dashed yellow lines denote the uni-directional attention.}
    \label{fig:basicmodel}
\end{figure*}

\subsection{Unlabeled Dataset: \textit{UnDial}}
Large clean dialogs are difficult to acquire. We build the unlabeled dialog corpora from various available sources, ranging from online forum chatting logs to customer service conversations. We select 14 existing dialog corpora and perform careful processing on all data. Then we acquire a large-scale unlabeled dialog dataset \textit{UnDial}, which consists of 35M utterances.
Table \ref{tab:pretraindial} shows the statistics of our final pre-training unlabeled data. For more details about the data statistics and the text processing method, please refer to Appendix A.2.



\section{Method}

\begin{table}[t]
    \centering
    \scalebox{0.9}{
\begin{tabular}{l|ccc}
\hline
Name        & \multicolumn{1}{c|}{\# Dialogs}                                                          & \multicolumn{1}{c|}{\# Utterance}                                                         & \# Unified DA                                                         \\ \hline
MultiWOZ    & \multicolumn{1}{c|}{10,433}                                                              & \multicolumn{1}{c|}{142,968}                                                              & 11                                                                    \\ \cline{2-4} 
Frames      & \multicolumn{1}{c|}{1,369}                                                               & \multicolumn{1}{c|}{19,986}                                                               & 14                                                                    \\ \cline{2-4} 
MSRe2e      & \multicolumn{1}{c|}{10,087}                                                              & \multicolumn{1}{c|}{74,686}                                                               & 12                                                                    \\ \cline{2-4} 
SGD         & \multicolumn{1}{c|}{22,825}                                                              & \multicolumn{1}{c|}{463,284}                                                              & 9                                                                     \\ \cline{2-4} 
DSTC2       & \multicolumn{1}{c|}{3,235}                                                               & \multicolumn{1}{c|}{44,532}                                                               & 7                                                                     \\ \cline{2-4} 
SimJoint    & \multicolumn{1}{c|}{3,008}                                                               & \multicolumn{1}{c|}{24,112}                                                               & 6                                                                     \\ \cline{2-4} 
STAR        & \multicolumn{1}{c|}{6,652}                                                               & \multicolumn{1}{c|}{107,846}                                                              & 11                                                                    \\ \cline{2-4} 
DailyDialog & \multicolumn{1}{c|}{13,117}                                                              & \multicolumn{1}{c|}{98,366}                                                               & 9                                                                     \\ \hline
UniDA       & \multicolumn{1}{c|}{70,726}                                                              & \multicolumn{1}{c|}{975,780}                                                              & 20                                                                    \\ \hline
Unified DAs & \multicolumn{3}{c}{\textit{\begin{tabular}[c]{@{}c@{}}request, select, reqalts, affirm, not\_sure, \\ inform, impl-confirm, expl-confirm,\\ notify\_success, notify\_failure, hi, bye, \\ negate, repeat, welcome, thank\_you, \\ direct, dont\_understand, propose, offer\end{tabular}}} \\ \hline
\end{tabular}}
    
    \caption{Statistics of the labeled dataset UniDA.}
    \label{tab:UniDA}
\end{table}

\begin{table}[t]
    \centering
    \scalebox{0.9}{
    \begin{tabular}{l|c}
\hline
\# Datasets             & 14    \\
\# Dialog Sessions     & 14M   \\
\# Utterances          & 35M  \\
Avg. Utterances per Dialog  & 2.5    \\
Avg. Tokens per Utterance   & 14.6   \\
\hline
\end{tabular}}
    
    \caption{Statistics of the unlabeled dataset UnDial.}
    \label{tab:pretraindial}
\end{table}

In this section, we first introduce the model architecture.
Then we describe each objective used in our pre-training and the proposed semi-supervised pre-training paradigm.



\subsection{Model Architecture}
We choose UniLM \cite{dong2019unified} as our backbone model.
It contains a bi-directional encoder for understanding and a uni-directional decoder for generation, which is  naturally suitable for task-oriented dialog modeling. The encoder and the decoder are weight-shared.
We adopt a similar scheme of input representation in \citet{bao2020plato}, where the input embeddings consist of four elements: tokens, roles, turns, and positions. 
Role embeddings are like segmentation embeddings in BERT and are used to differentiate which role the current token belongs to, either user or system. Turn embeddings are assigned to each token according to its turn number. Position embeddings are assigned to each token according to its relative position within its belonging sentence. More details can be found in Appendix B.1.

\subsection{Pre-training Objectives}
Four objectives are employed in our dialog pre-training process:
response selection, response generation,  DA prediction and consistency regularization. 
Figure \ref{fig:basicmodel} illustrates the procedure of pre-training. 

\paragraph{Response Selection.} Many work \cite{wu2020tod, bao2020plato, henderson2019convert} show that the response selection task can capture the coherency between dialog contexts and responses and thus benefit dialog understanding. 
We follow their implementation and model this task as a binary classification problem. 
Specifically, for a context response pair $(c, r)$ from the corpus,
the positive example (with label $l=1$) is obtained by concatenating $c$ with its corresponding response $r$,
and the negative example (with label $l=0$) is constructed by concatenating $c$ with a response $r^-$ that is randomly selected from the corpus.
A binary cross-entropy loss is defined as:
\begin{equation}
    \mathcal{L}_{\mathrm{RS}}=-\log p \left(l=1| c, r\right) - \log p\left(l=0| c, r^-\right)
\end{equation}
in which the classification probability $p \left(l| c, r\right)$ is calculated by feeding the concatenated sequence of $c$ and $r$ into the bi-directional encoder and adding a binary classification head on the extracted representation $h_{cls}$ of token [CLS] from the last transformer layer:
\begin{equation}
    p\left(l=1| c, r \right) = \operatorname{sigmoid}\left(\phi_a (h_{cls})  \right) \in \mathbb{R}^1
\end{equation}
where $\phi_a$ is a fully-connected neural network with the output layer of size 1. $\operatorname{sigmoid}$ is the sigmoid function acts on each dimension of the input vector.


\paragraph{Response Generation.}
The response generation task aims to predict the dialog response $r$ auto-regressively based on the dialog context $c$.
We adopt the standard negative log-likelihood loss for the generation task:
\begin{equation}
\mathcal{L}_{\mathrm{RG}} =-\sum_{t=1}^{T} \log 
p\left(r_t|c, r_{<t}\right)
\end{equation}
where $r_t$ is the $t$-th word in $r$, $r_{<t}=\{r_1, ..., r_{t-1}\}$ represents the  words of previous steps.


\paragraph{DA Prediction.}

For a context response pair $(c, r)$ sampled from UniDA, the DA prediction task aims to predict the DA label $a$ of the response $r$ based merely on the context $c$. 
Note that, since there are some responses in UniDA are associated with multiple DAs, we model the DA prediction task as a multi-label classification problem. 
We denote $a=(a_1, a_2,...,a_N)$, where $N$ is the total number of dialog acts. A multi-dimensional Bernoulli  distribution is used for dialog acts: $p(a|c)=\prod_i^{N} p(a_i|c)$.  
Taking the dialog context $c$ as input, we add a multi-dimensional binary classifiers on $h_{cls}$ to predict each act $a_i$. The binary classification loss is:

\begin{equation}
\begin{aligned}
\mathcal{L}_{\mathrm{DA}} 
    =-\sum_{i=1}^{N} &\left\{ y_i\log p(a_i| c)\right. \\
    &+\left. (1-y_i)\log \left(1-p(a_i| c)\right)\right\}
\end{aligned}
\end{equation}

\begin{equation}
\label{act-recog-mapping}
    p\left(a|c\right)=\operatorname{sigmoid}\left(\phi_{b} \left(h_{cls}\right) \right) \in \mathbb{R}^{N}
\end{equation}
where $\phi_b$ is a fully-connected neural network with the output layer of size $N$. 
$y_i\in\{0,1\}$ is the true label of $a_i$.

\begin{figure}[t]
    \centering
    \includegraphics[width=0.25\textwidth]{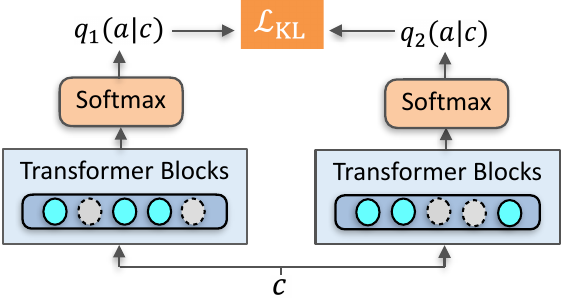}
    \caption{The procedure of computing $\mathcal{L}_\mathrm{KL}$.}
    \label{fig:gating}
\end{figure}

\paragraph{Consistency Regularization.}
For UnDial, the DA annotations are unavailable. In that case, we need to infer the DA labels based on the  given dialog context $c$. Instead of using $p(a|c)$ in Eq. (\ref{act-recog-mapping}), we use a categorical distribution $q(a|c)$ for dialog acts:
\begin{equation}
    q\left(a|c\right)=\operatorname{softmax}\left(\phi_{b} (h_{cls})\right) \in \mathbb{R}^{N}
\end{equation}
where $\operatorname{softmax}$ is the softmax function, $\phi_b$ is the same feed-forward neural network in Eq. (\ref{act-recog-mapping}). So $\sum_{i=1}^Nq(a_i|c)=1$. 
Then we employ a dropout-based consistency regularization to learn better representations \cite{gao2021simcse}. 
Concretely, given the same dialog context $c$, we feed $c$ to go through the forward pass of the model twice. Due to the randomness of the dropout mechanism in transformers, we can get two different sets of hidden features, and therefore, two different categorical distributions of dialog policy, denoted as $q_1(a|c)$ and $q_2(a|c)$.
Then the Kullback-Leibler (KL) divergence between these two output distributions is calculated as $\mathcal{D}_{KL}(q_1\|q_2)$.
We minimize the bidirectional KL divergence as in \cite{liang2021r} between the two distributions to regularize the model predictions, which is defined as:
\begin{equation}
\mathcal{L}_{KL}=\frac{1}{2}\left(
\mathcal{D}_{KL}\left(q_{1} \| q_{2}\right)+ 
\mathcal{D}_{\mathrm{KL}}\left(q_{2} \| q_{1}\right)
\right)
\end{equation}
Figure \ref{fig:gating} illustrate the procedure of computing $\mathcal{D}_{KL}$.

\subsection{Semi-supervised Pre-training Paradigm}

We aim to leverage semi-supervised pre-training to learn better pre-trained representations from both the labeled and unlabeled data. For the labeled dataset UniDA, we use all objectives to optimize. 
The total loss $\mathcal{L}_\mathrm{label}$ is computed as:
\begin{equation}
\mathcal{L}_\mathrm{label}=\mathcal{L}_\mathrm{RS} + \mathcal{L}_\mathrm{RG} + \mathcal{L}_\mathrm{DA}+  \mathcal{L}_\mathrm{KL}
\label{label_loss}
\end{equation}

For the unlabeled data UnDial, since some dialogs collected from the open-domain Internet are too noisy to be compatible with our DA taxonomy, we propose to use a gating mechanism to select a high-quality subset of UnDial for prediction.
In practice, we compute a soft gating score $g\in [0,1]$ based on the entropy of $q(a|c)$ to control whether a data point is adopted for consistency regularization in the current iteration. 
\begin{equation}
    g= \min\left\{\max\left\{0, \frac{E_{max} - (E + \log E)}{E_{max}}\right\}, 1\right\}
\end{equation}
where $E_{max}=\log N$ is the Maximum Entropy of $N$-dimensional probability distribution. $E$ is the current entropy of $q(a|c)$, i.e., $E=\sum_i^N q(a_i|c) \log  q(a_i|c)$. In practice, we use the perturbed distribution $q_1(a|c)$ as the approximation of $q(a|c)$ to calculate the gate score. 

Hence, we have the loss $\mathcal{L}_\mathrm{unlabel}$ for the unlabeled data to adjust it adaptively by the gate $g$ as following:
\begin{equation}
    \mathcal{L}_\mathrm{unlabel}=\mathcal{L}_\mathrm{RS} +  \mathcal{L}_\mathrm{RG} + g \mathcal{L}_\mathrm{KL} 
\end{equation}


The final loss $\mathcal{L}_\mathrm{pre}$ is computed as:
\begin{equation}
\mathcal{L}_\mathrm{pre}=\mathcal{L}_\mathrm{unlabel} + \mathcal{L}_\mathrm{label}
\end{equation}

In the pre-training process, we mix and shuffle UniDA and UnDial, and randomly sample batches from the mixed corpus.



\subsection{Fine-tuning and Inference}
In the fine-tuning stage, we concentrate on task-oriented dialog tasks. For tasks that contained necessary semantic labels (e.g., belief states and dialog acts), we re-organize the response $r$ to contain those labels, and generate them together. Suppose the sequence of the labels is $d$. Thus the new response $r^{*}=(d,r)$ is the concatenation of $d$ and $r$ and is generated in the downstream tasks. For tasks that do not have semantic labels, we generate the initial response $r$.
We also maintain the DA prediction task to alleviate the model discrepancy between pre-training and fine-tuning \cite{zenginvestigation}. Therefore, The fine-tuning loss is as follows:
\begin{equation}
\mathcal{L}_\mathrm{fine}=\mathcal{L}_\mathrm{RS} + \mathcal{L}_\mathrm{RG} + \alpha\mathcal{L}_\mathrm{DA}
\end{equation}
where $\alpha=1$ for tasks that provide DA annotations and $\alpha=0$ for tasks that contain no DA annotations.

\begin{table*}[t]
\centering
\scalebox{0.8}{
\begin{tabular}{l|cccc|cccc}
\hline
\multirow{2}{*}{Model} & \multicolumn{4}{c|}{MultiWOZ2.0} & \multicolumn{4}{c}{MultiWOZ2.1} \\ 
 & \texttt{Inform} & \texttt{Success} & \texttt{BLEU} & \texttt{Comb} & \texttt{Inform} & \texttt{Success} & \texttt{BLEU} & \texttt{Comb} \\ \hline
 
SimpleTOD \cite{hosseini2020simple} & 84.40 & 70.10 & 15.01 & \multicolumn{1}{c|}{92.26} & 85.00 & 70.50 & 15.23 & 92.98 \\
DoTS \cite{jeon2021domain} & 86.59 & 74.14 & 15.06 & \multicolumn{1}{c|}{95.43} & 86.65 & 74.18 & 15.90 & 96.32 \\
SOLOIST \cite{peng2020soloist} & 85.50 & 72.90 & 16.54 & \multicolumn{1}{c|}{95.74} & -- & -- & -- & -- \\
MinTL \cite{lin2020mintl} & 84.88 & 74.91 & 17.89 & \multicolumn{1}{c|}{97.79} & -- & -- & -- & -- \\
PPTOD \cite{su2021multitask} & 89.20 & 79.40 & 18.62 & \multicolumn{1}{c|}{102.92} & 87.09 & 79.08 & 19.17 & 102.26 \\
UBAR \cite{yang2020ubar} & \textbf{95.40} & 80.70 & 17.00 & \multicolumn{1}{c|}{105.05} & \textbf{95.70} & 81.80 & 16.50 & 105.25 \\
GALAXY(w/o pre-train) & 93.10 & 81.00 & 18.44 & \multicolumn{1}{c|}{105.49} & 93.50 & 81.70 & 18.32 & 105.92 \\
GALAXY & 94.40 & \textbf{85.30} & \textbf{20.50} & \multicolumn{1}{c|}{\textbf{110.35}} & 95.30 & \textbf{86.20} & \textbf{20.01} & \textbf{110.76} \\ \hline
\end{tabular}
}
\caption{E2E performances on MultiWOZ2.0/2.1. All results are from original papers. `w/o pre-train' means using original weights of UniLM for initialization.}
\label{multiwoz}
\vspace{-0.5cm}
\end{table*}

\begin{table}[t]
\centering
\scalebox{0.8}{
\begin{tabular}{l|cccc}
\hline
Model & \texttt{Match} & \texttt{SuccF1} & \texttt{BLEU} & \texttt{Comb}\\ \hline
SEDST \cite{jin2018explicit} & 84.50 & 82.90 & 19.30  & 103.00 \\
TSCP \cite{lei2018sequicity}& 84.50 & 81.10 & 21.90 & 104.70 \\
LABES \cite{zhang-etal-2020-probabilistic} & \textbf{85.80} & 77.00 & 22.80& 104.20 \\
FSDM \cite{shu2019flexibly}& 84.80 & 82.10 & 21.50 & 104.95 \\
GALAXY (w/o pre-train) & 81.90 & 83.30 & 22.00 & 104.60 \\
GALAXY & 85.30 & \textbf{83.60} & \textbf{23.00} & \textbf{107.45}\\ \hline
\end{tabular}}
\caption{E2E performances on In-Car. All results are from original papers. `w/o pre-train' means using original weights of UniLM for initialization.}
\label{tab: incar}
\vspace{-0.5cm}
\end{table}

\section{Experimental Settings}
\subsection{Evaluation Datasets}
We evaluate the end-to-end dialog system performance of GALAXY on two well-studied task-oriented dialog benchmarks: Stanford In-Car Assistant (In-Car) \cite{eric2017key},  
MultiWOZ \cite{budzianowski2018multiwoz}.
In-Car consists of dialogs between a user
and an in-car assistant system covering three tasks: calendar scheduling, weather information retrieval, and point-of-interest navigation.
Following the data processing in \cite{zhang-etal-2020-probabilistic}, we divide the dataset into training/validation/testing sets with 2425/302/304 dialogs respectively.
MultiWOZ is a large-scale human-human dataset spanning seven domains, which is one of the most challenging datasets in task-oriented dialog due to its complex ontology and diverse language styles.
We evaluate our model on MultiWOZ2.0 (the original version) and MultiWOZ2.1 (a revised version) since both are popular benchmarks with various competing models.
Following the data processing in \citet{yang2020ubar}, we obtain 8438/1000/1000 dialogs for training/validation/testing respectively. We also adopt delexicalized responses for task-oriented generation, which allows the model to learn value-independent parameters \cite{zhang2020task}.


\subsection{Evaluation Metrics}
We use \texttt{BLEU} \cite{papineni2002bleu} to measure the response generation quality. Metrics relate to task completion are used for separate datasets to facilitate comparison
with prior works. For MultiWOZ, we report \texttt{Inform}, \texttt{Success},  
as a combined score (\texttt{Comb}) is also computed via (\texttt{Inform} $+$ \texttt{Success})$\times0.5 + $\texttt{BLEU}
as an overall quality measure 
as in \citet{mehri2019structured}. For In-Car, we use \texttt{Match} and \texttt{SuccF1} following \citet{lei2018sequicity}, and calculate a similar combined score (\texttt{Comb}) via (\texttt{Match} $+$ \texttt{SuccF1})$\times0.5 + $\texttt{BLEU}.




\section{Experimental Results}

In our experiments, we focus on the setting of end-to-end dialog modeling (E2E), in which no ground-truth immediate labels are provided to the model. GALAXY is initialized with UniLM and then performs semi-supervised pre-training with UniDA and UnDial. Notably, we removed the validation and testing set of MultiWOZ from UniDA during pre-training for fairness. We compare GALAXY with all published work on respective datasets. We also compare different pre-trained conversation models (PCMs) and different semi-supervised pre-training methods to verify the efficacy of GALAXY. In addition, we conduct an extensive discussion and analysis to reveal the internal performance of GALAXY. More details about implementation can be found in Appendix B.2.

\subsection{Benchmark Performance}
As shown in Table \ref{multiwoz} and Table \ref{tab: incar}, GALAXY achieves new state-of-the-art combined scores on all datasets, improving In-Car by 2.5 points (from 104.95 to 107.45), MultiWOZ2.0 by 5.3 points (from 105.05 to 110.35), and MultiWOZ2.1 by 5.5 points (from  105.25 to 110.76).
Note that in both tables, GALAXY is the only model that can obtain best \texttt{Success} while maintaining \texttt{BLEU} at a very high level, which means that GALAXY can take better dialog policy than other models to facilitate task completion, and therefore generate better responses. Our model can also achieve competitive results in \texttt{Inform} on par with other best baselines.
We also report the results of GALAXY (w/o pre-train) without the pre-training procedure on more dialog corpora. From both tables, GALAXY also achieves comparable results with previous best models, indicating that our model architecture is competitive for dialog modeling. More E2E results given oracle belief states on MultiWOZ are shown in Appendix D.

\subsection{Comparison with Other PCMs}
We verify that GALAXY has a much better ability to fulfill task-oriented dialog tasks than other PCMs due to modeling dialog policy during pre-training. To alleviate the discrepancy brought from model structure, we use UniLM \cite{dong2019unified} and PLATO \cite{bao2020plato} as our baselines. We also train both models on our pre-training dialog datasets (UniDA and UnDial) with their original objectives and perform the same fine-tuning process on MultiWOZ2.0. We denote the new models as TOD-UniLM and TOD-PLATO, respectively.
As shown in Table \ref{PCM_compare}, the results of both models are worse than GALAXY due to the lack of using important information of dialog policy.

\subsection{Comparison with Other Semi-supervised Pre-training Methods}
As shown in Table \ref{SSL_compare}, we also compare GALAXY with other semi-supervised pre-training methods on MultiWOZ2.0. Specifically, we employ three baselines: Pseudo-Labeling, Variation Autoencoder (VAE), and multi-task learning. More details about the first two approaches are offered in Appendix C. 
For multi-task learning, we discard the $\mathcal{L}_\mathrm{KL}$ loss for GALAXY, which represents that model does not perform any inference over DA labels on UnDial. We denote this method as GALAXY$_{multi}$. The results in Table \ref{SSL_compare} show that VAE has the worst performance because it is difficult to pre-train stochastic latent variables well.
Multi-task learning is the most substantial baseline among the three methods, which indicates the importance of integrating DA annotations in the pre-training process.  However, without inference on unlabeled dialog samples, GALAXY$_{multi}$ can not explore the stored knowledge of dialog policy thoroughly.

\begin{table}[t]
\centering
\scalebox{0.8}{
\begin{tabular}{lcccc}
\hline
Model & \texttt{Inform} & \texttt{Success} & \texttt{BLEU} & \texttt{Comb} \\ \hline
UniLM & 92.40 &81.40	&18.45&	105.35\\
PLATO & 91.20 & 77.20 & 16.68 & 100.88 \\
TOD-UniLM & 93.50 & 81.30 & 19.13 & 106.53 \\
TOD-PLATO & 92.10 & 79.40 & 17.23 & 102.98 \\
GALAXY & \textbf{94.40} & \textbf{85.30} & \textbf{20.50} & \textbf{110.35} \\ \hline
\end{tabular}}
\caption{E2E performances of different pre-trained conversation models on MultiWOZ2.0.}
\label{PCM_compare}
\end{table}

\begin{table}[t]
\centering
\scalebox{0.8}{
\begin{tabular}{lcccc}
\hline
Model & \texttt{Inform} & \texttt{Success} & \texttt{BLEU} & \texttt{Comb} \\ \hline
Pseudo-Labeling & 90.10 & 80.30 & 16.79 & 101.99 \\
VAE & 89.00 & 76.40 & 16.48 & 99.18 \\
GALAXY$_{multi}$ & 93.90 & 82.30 & 19.17 & 107.27 \\
GALAXY & \textbf{94.40} & \textbf{85.30} & \textbf{20.50} & \textbf{110.35} \\ \hline
\end{tabular}}
\caption{E2E performance of different semi-supervised pre-training methods on MultiWOZ2.0.}
\label{SSL_compare}
\vspace{-0.5cm}
\end{table}

\begin{table*}[htp]
\scalebox{0.8}{
\begin{tabular}{l|ccc|ccc|ccc|ccc}
\hline
\multirow{2}{*}{Model} & \multicolumn{3}{c|}{5\% data} & \multicolumn{3}{c|}{10\% data} & \multicolumn{3}{c|}{20\% data} & \multicolumn{3}{c}{50\% data} \\
 & \texttt{Inform} & \texttt{Success} & \texttt{BLEU} & \texttt{Inform} & \texttt{Success} & \texttt{BLEU} & \texttt{Inform} & \texttt{Success} & \texttt{BLEU} & \texttt{Inform} & \texttt{Success} & \texttt{BLEU} \\ \hline
DAMD & 56.60 & 24.50 & 10.60 & 62.00 & 39.40 & 14.50 & 77.90 & 70.30 & 12.10 & 83.00 & 72.90 & 16.90 \\
SOLOIST & 69.30 &52.30& 11.80& 69.90 & 51.90& 14.60& 74.00 &60.10 &15.24 & -- & --& -- \\ 
MinTL & 75.48 & 60.96 & 13.98 & 78.08 & 66.87 & 15.46 & 82.48 & 68.57 & 13.00 & 90.10$^{*}$ & 78.60$^{*}$ & 17.90$^{*}$ \\
PPTOD & 79.86 & 63.48 & 14.89 & 84.42 & 68.36 & 15.57 & 84.94 & 71.70 & 17.01 & -- & --& -- \\
UBAR & 73.04$^{*}$ & 60.28$^{*}$ & 16.03$^{*}$ & 79.20$^{*}$ & 68.70$^{*}$ & 16.09$^{*}$ & 82.50$^{*}$ & 66.60$^{*}$ & \textbf{17.72}$^{*}$ & 91.50$^{*}$ & 78.20$^{*}$ & 17.05$^{*}$ \\
GALAXY & \textbf{80.59} & \textbf{67.43} & \textbf{17.39} & \textbf{87.00} & \textbf{75.00} & \textbf{17.65} & \textbf{89.55} & \textbf{75.85} & 17.54 & \textbf{93.35} & \textbf{82.35} & \textbf{18.37} \\ \hline
\end{tabular}}
\centering
\caption{E2E results of low-resource experiments. 5\% (400 dialogs), 10\% (800 dialogs), 20\% (1600 dialogs), 50\% (4000 dialogs) of training data is used to train each model. $*$ denotes our re-implementation results.}
\label{fewshot}
\vspace{-0.1cm}
\end{table*}

\begin{table}[htp]
\centering
\scalebox{0.77}{
\begin{tabular}{lcccc}
\hline
Model & \texttt{Inform} & \texttt{Success} & \texttt{BLEU} & \texttt{Comb} \\ \hline
GALAXY & \textbf{94.40} & \textbf{85.30} & \textbf{20.50} & \textbf{110.35} \\ 
 \ \ \ $-g$ & 94.20 & 83.50 & 19.26 & 108.11 \\
\ \ \ $-\mathcal{L}_\mathrm{DA}$ & 89.10 & 79.90 & 18.77 & 103.27 \\
\ \ \  $-\mathcal{L}_\mathrm{KL}$ & 93.90 & 82.30 & 19.17 & 107.27 \\
\ \ \ $-\mathcal{L}_\mathrm{DA} - \mathcal{L}_\mathrm{KL}$ & 93.30 & 81.20 & 19.54 & 106.79 \\

\hline
\end{tabular}}
\caption{E2E results of ablation study on MultiWOZ2.0.}
\label{ablation}
\end{table}

\begin{figure}[htp]
    \centering
    \includegraphics[width=0.43\textwidth]{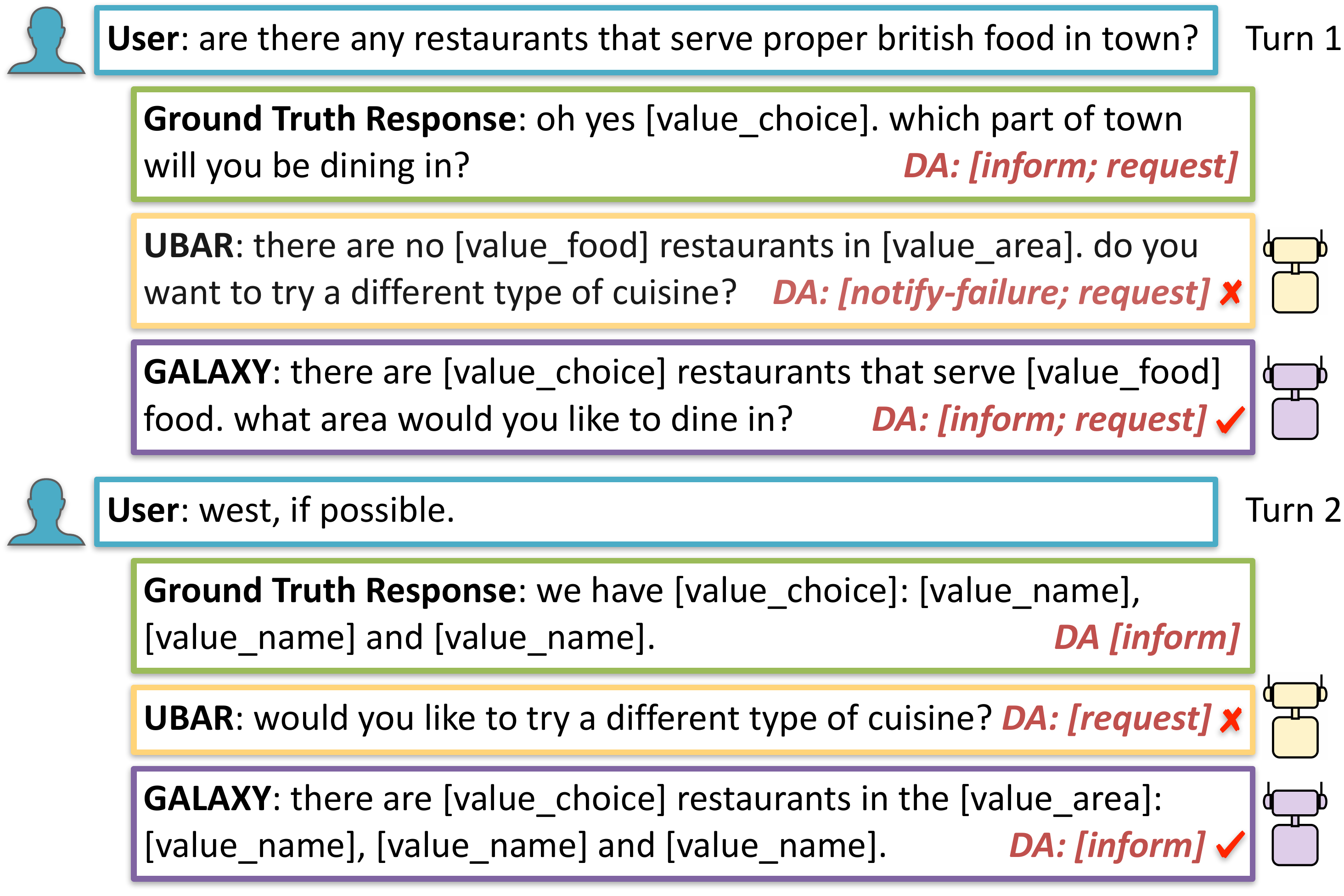}
    \vspace{-0.2cm}
    \caption{Case Study: Delexicalized responses generated by GALAXY and UBAR on MultiWOZ2.0 test data. 
    }
    \label{fig:case_study}
    \vspace{-0.2cm}
\end{figure}

\subsection{Low Resource Evaluation}
Many recent works \cite{peng2020few,wu2020tod} have demonstrated that pre-trained models have a solid few-shot ability in the understanding and conditional generation tasks. We also evaluate GALAXY in the simulated low resource setting on MultiWOZ2.0, showing that it is more sample-efficiency than existing models. Specifically, we use 5\%, 10\%, 20\%, and 50\% of the training set data to train our models and baselines.
To be fair, we discard the (1-X\%) training data of MultiWOZ from UniDA in the pre-training process under each X\% setting, eliminating the influence of using any external data.
Compared baselines include: DAMD \cite{zhang2020task}, SOLOIST \cite{peng2020soloist}, MinTL \cite{lin2020mintl}, PPTOD \cite{su2021multitask} and UBAR \cite{yang2020ubar}. Experimental results in Table \ref{fewshot} show that GALAXY significantly outperforms other models under all low-resource settings.

\section{Analysis and Discussion}
In this section, we try to answer three questions: (1) How does our semi-supervised method work during the pre-training process? (2) How much improvements does $\mathcal{L}_\mathrm{DA}$,    $\mathcal{L}_\mathrm{KL}$ and the gating mechanism contribute? 
(3) How can our model improve task completion in real cases?

\begin{figure}[t]
    \centering
    \includegraphics[width=0.35\textwidth]{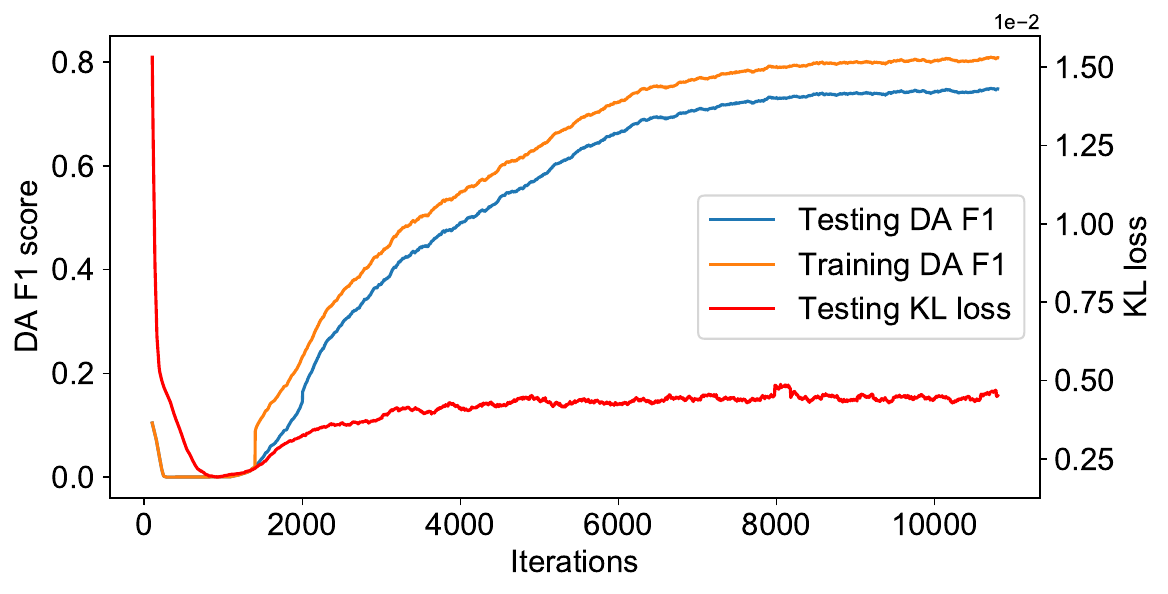}
    \vspace{-0.3cm}
    \caption{Learning curves of  train/test DA F1 scores and the  $\mathcal{L}_\mathrm{KL}$ loss.}
    \label{fig:curve}
\end{figure}

\paragraph{Learning Curve.} In order to figure out how consistency regularization loss can influence the pre-training, we monitor the predicted DA accuracy and $\mathcal{L}_\mathrm{KL}$. Specifically, we conduct a simulated experiment where 10\% UniDA and 100\% UnDial are used for training, and the rest of UniDA is held out as a testing set. Then we observe the testing DA F1 score and the $\mathcal{L}_\mathrm{KL}$ loss on the rest of UniDA data. Note that our goal is to mimic the actual case that whether the model can learn well given limited labeled data and large unlabeled data.
As we can see from Figure \ref{fig:curve},  $\mathcal{L}_\mathrm{KL}$ decreases to zero at the beginning, indicating that the model falls into the \textit{collapsing} mode \cite{chen2021exploring}, which means all outputs collapse to a constant. 
However, since we have the $\mathcal{L}_\mathrm{DA}$ loss on labeled data, the collapsing problem can be tackled in the following iterations. 
On the other hand, the regularization loss $\mathcal{L}_\mathrm{KL}$ performs on the labeled data can also avoid over-fitting to some extent, which is shown in Figure \ref{fig:curve} that the testing DA F1 score keeps increasing during the pre-training without degradation.


\paragraph{Ablation Results.} 
Table \ref{ablation} shows the ablation results of GALAXY on MultiWOZ2.0. Without $\mathcal{L}_\mathrm{DA}$, GALAXY performs worst because of the collapsing problem. GALAXY without $\mathcal{L}_\mathrm{KL}$ equals to multi-task learning, but the results are not as good as our semi-supervised learning due to the inadequate utilization of unlabeled data. If we discard both losses, which backs to the use of common pre-training objectives $\mathcal{L}_\mathrm{RS}$ and $\mathcal{L}_\mathrm{RG}$, we can acquire 106.79 in \texttt{Comb}, suggesting that our pre-training dialog datasets are high-quality and can facilitate task-oriented dialog training. We also examine the function of the gating mechanism.  Note that adding the gate $g$ is essential for improving model performance (\texttt{Comb} increase from 108.11 to 110.35), indicating that it can filter inappropriate data for our semi-supervised pre-training. 
Table \ref{tab:gate} shows the predicted gating scores of four utterances from UnDial and the DAs annotated manually for the corresponding responses.

\paragraph{Case Study.} Figure \ref{fig:case_study} illustrates a case where GALAXY chooses correct dialog acts for the first two turns so that the whole conversation can steer towards successful task completion. On the contrary, UBAR takes a wrong DA \textit{notify-failure} at the beginning turn and a redundant DA \textit{request} at the second turn, which leads to a failure for the interaction.



\begin{table}[t]
\centering
\scalebox{0.75}{
\begin{tabular}{ll}
\hline
Context   & i need either the email address , or just zip code. \textbf{(Gate: 1.0)} \\
Response & zip code : 24627.  \textbf{(DA: inform)}                                \\ \hline
Context   & i need to return an item , can you help me? \textbf{(Gate: 0.91)}        \\
Response & sure , may i have your name please?         \textbf{(DA: request)}     \\ \hline
Context   & i pour a little liquor out for habeas. \textbf{(Gate: 0.41)}             \\
Response & i pour it into corpus. \textbf{(DA: N.A.)}                               \\ \hline
Context   & one word : justice. \textbf{(Gate: 0.19)}                                \\
Response & let me guess , you drive a 1980 ford pinto. \textbf{(DA: N.A.)}          \\ \hline
\end{tabular}}
\caption{ Examples of predicted gating scores give the context. Responses are also annotated with DAs  for analysis. `N.A.' means we cannot find a suitable DA for the response.}
\label{tab:gate}
\vspace{-0.5cm}
\end{table}

\section{Conclusion}
In this paper, we propose GALAXY, a  pre-trained conversation model that learns dialog policy explicitly in the pre-training process via semi-supervised learning.
We introduce a  dialog  act  prediction task for policy optimization
and use a consistency regularization loss to learn better representations on unlabeled dialog corpora. 
A gating mechanism is also used to weigh suitable unlabeled samples. 
Experiments show that our model creates new SOTA results on several task-oriented dialog benchmarks and outperforms existing models by a large margin in various low-resource settings.
We hope that GALAXY, and the newly collected labeled dataset \textit{UniDA} and large-scale unlabeled corpus \textit{UnDial},  can inspire researchers  to explore the new paradigm to build pre-trained conversation models for task-oriented dialog.


\newpage
\section*{Acknowledgement}
This work was supported by Alibaba Group through Alibaba Research Intern Program.
This work was partially supported by National Natural Science Foundation of China (No. 61906185), Youth Innovation Promotion Association of CAS China (No. 2020357), Shenzhen Science and Technology Innovation Program (Grant No. KQTD20190929172835662), Shenzhen Basic Research Foundation (No. JCYJ20200109113441941).
We also thank Dr. Yichi Zhang for the cute cartoon in Figure 1.

\bibliography{pcm}

\begin{thebibliography}{72}
\providecommand{\natexlab}[1]{#1}

\bibitem[{Adiwardana et~al.(2020)Adiwardana, Luong, So, Hall, Fiedel,
  Thoppilan, Yang, Kulshreshtha, Nemade, Lu et~al.}]{adiwardana2020towards}
Adiwardana, D.; Luong, M.-T.; So, D.~R.; Hall, J.; Fiedel, N.; Thoppilan, R.;
  Yang, Z.; Kulshreshtha, A.; Nemade, G.; Lu, Y.; et~al. 2020.
\newblock Towards a human-like open-domain chatbot.
\newblock \emph{arXiv preprint arXiv:2001.09977}.

\bibitem[{Asri et~al.(2017)Asri, Schulz, Sharma, Zumer, Harris, Fine, Mehrotra,
  and Suleman}]{asri2017frames}
Asri, L.~E.; Schulz, H.; Sharma, S.; Zumer, J.; Harris, J.; Fine, E.; Mehrotra,
  R.; and Suleman, K. 2017.
\newblock Frames: a corpus for adding memory to goal-oriented dialogue systems.
\newblock \emph{arXiv preprint arXiv:1704.00057}.

\bibitem[{Bao et~al.(2020)Bao, He, Wang, Wu, and Wang}]{bao2020plato}
Bao, S.; He, H.; Wang, F.; Wu, H.; and Wang, H. 2020.
\newblock {PLATO}: Pre-trained Dialogue Generation Model with Discrete Latent
  Variable.
\newblock In \emph{Proceedings of the 58th Annual Meeting of the Association
  for Computational Linguistics}. Online: Association for Computational
  Linguistics.

\bibitem[{Budzianowski and Vuli{\'c}(2019)}]{budzianowski2019hello}
Budzianowski, P.; and Vuli{\'c}, I. 2019.
\newblock Hello, it's GPT-2--how can I help you? towards the use of pretrained
  language models for task-oriented dialogue systems.
\newblock \emph{arXiv preprint arXiv:1907.05774}.

\bibitem[{Budzianowski et~al.(2018)Budzianowski, Wen, Tseng, Casanueva, Ultes,
  Ramadan, and Ga{\v{s}}i{\'c}}]{budzianowski2018multiwoz}
Budzianowski, P.; Wen, T.-H.; Tseng, B.-H.; Casanueva, I.; Ultes, S.; Ramadan,
  O.; and Ga{\v{s}}i{\'c}, M. 2018.
\newblock MultiWOZ--A Large-Scale Multi-Domain Wizard-of-Oz Dataset for
  Task-Oriented Dialogue Modelling.
\newblock \emph{arXiv preprint arXiv:1810.00278}.

\bibitem[{Bunt(2009)}]{bunt2009dit}
Bunt, H. 2009.
\newblock The DIT++ taxonomy for functional dialogue markup.
\newblock In \emph{AAMAS 2009 Workshop, Towards a Standard Markup Language for
  Embodied Dialogue Acts}, 13--24.

\bibitem[{Bunt et~al.(2010)Bunt, Alexandersson, Carletta, Choe, Fang, Hasida,
  Lee, Petukhova, Popescu-Belis, Romary et~al.}]{bunt2010towards}
Bunt, H.; Alexandersson, J.; Carletta, J.; Choe, J.-W.; Fang, A.~C.; Hasida,
  K.; Lee, K.; Petukhova, V.; Popescu-Belis, A.; Romary, L.; et~al. 2010.
\newblock Towards an ISO standard for dialogue act annotation.
\newblock In \emph{Seventh conference on International Language Resources and
  Evaluation (LREC'10)}.

\bibitem[{Byrne et~al.(2019)Byrne, Krishnamoorthi, Sankar, Neelakantan,
  Duckworth, Yavuz, Goodrich, Dubey, Kim, and Cedilnik}]{Taskmaster}
Byrne, B.; Krishnamoorthi, K.; Sankar, C.; Neelakantan, A.; Duckworth, D.;
  Yavuz, S.; Goodrich, B.; Dubey, A.; Kim, K.-Y.; and Cedilnik, A. 2019.
\newblock Taskmaster-1: Toward a Realistic and Diverse Dialog Dataset.

\bibitem[{Chen et~al.(2021)Chen, Chen, Yang, Lin, and Yu}]{chen2021action}
Chen, D.; Chen, H.; Yang, Y.; Lin, A.; and Yu, Z. 2021.
\newblock Action-Based Conversations Dataset: A Corpus for Building More
  In-Depth Task-Oriented Dialogue Systems.
\newblock \emph{arXiv preprint arXiv:2104.00783}.

\bibitem[{Chen and He(2021)}]{chen2021exploring}
Chen, X.; and He, K. 2021.
\newblock Exploring simple siamese representation learning.
\newblock In \emph{CVPR}, 15750--15758.

\bibitem[{Dai et~al.(2021)Dai, Li, Li, Sun, Huang, Si, and
  Zhu}]{dai-etal-2021-preview}
Dai, Y.; Li, H.; Li, Y.; Sun, J.; Huang, F.; Si, L.; and Zhu, X. 2021.
\newblock Preview, Attend and Review: Schema-Aware Curriculum Learning for
  Multi-Domain Dialogue State Tracking.
\newblock In \emph{ACL-IJCNLP}, 879--885.

\bibitem[{Danescu-Niculescu-Mizil and
  Lee(2011)}]{danescu-niculescu-mizil-lee-2011-chameleons}
Danescu-Niculescu-Mizil, C.; and Lee, L. 2011.
\newblock Chameleons in Imagined Conversations: A New Approach to Understanding
  Coordination of Linguistic Style in Dialogs.
\newblock In \emph{Proceedings of the 2nd Workshop on Cognitive Modeling and
  Computational Linguistics}, 76--87. Portland, Oregon, USA: Association for
  Computational Linguistics.

\bibitem[{Devlin et~al.(2018)Devlin, Chang, Lee, and
  Toutanova}]{devlin2018bert}
Devlin, J.; Chang, M.-W.; Lee, K.; and Toutanova, K. 2018.
\newblock Bert: Pre-training of deep bidirectional transformers for language
  understanding.
\newblock \emph{arXiv preprint arXiv:1810.04805}.

\bibitem[{Dong et~al.(2019)Dong, Yang, Wang, Wei, Liu, Wang, Gao, Zhou, and
  Hon}]{dong2019unified}
Dong, L.; Yang, N.; Wang, W.; Wei, F.; Liu, X.; Wang, Y.; Gao, J.; Zhou, M.;
  and Hon, H.-W. 2019.
\newblock Unified Language Model Pre-training for Natural Language
  Understanding and Generation.
\newblock In \emph{33rd Conference on Neural Information Processing Systems
  (NeurIPS 2019)}.

\bibitem[{Eric and Manning(2017)}]{eric2017key}
Eric, M.; and Manning, C.~D. 2017.
\newblock Key-value retrieval networks for task-oriented dialogue.
\newblock \emph{arXiv preprint arXiv:1705.05414}.

\bibitem[{Fainberg et~al.(2018)Fainberg, Krause, Dobre, Damonte, Kahembwe,
  Duma, Webber, and Fancellu}]{fainberg2018talking}
Fainberg, J.; Krause, B.; Dobre, M.; Damonte, M.; Kahembwe, E.; Duma, D.;
  Webber, B.; and Fancellu, F. 2018.
\newblock Talking to myself: self-dialogues as data for conversational agents.
\newblock \emph{arXiv preprint arXiv:1809.06641}.

\bibitem[{Gao, Yao, and Chen(2021)}]{gao2021simcse}
Gao, T.; Yao, X.; and Chen, D. 2021.
\newblock SimCSE: Simple Contrastive Learning of Sentence Embeddings.
\newblock \emph{arXiv preprint arXiv:2104.08821}.

\bibitem[{Goodfellow et~al.(2014)Goodfellow, Pouget-Abadie, Mirza, Xu,
  Warde-Farley, Ozair, Courville, and Bengio}]{goodfellow2014generative}
Goodfellow, I.; Pouget-Abadie, J.; Mirza, M.; Xu, B.; Warde-Farley, D.; Ozair,
  S.; Courville, A.; and Bengio, Y. 2014.
\newblock Generative adversarial nets.
\newblock \emph{Advances in neural information processing systems}, 27.

\bibitem[{Gopalakrishnan et~al.(2019)Gopalakrishnan, Hedayatnia, Chen,
  Gottardi, and Hakkani-Tür}]{2019Topical}
Gopalakrishnan, K.; Hedayatnia, B.; Chen, Q.; Gottardi, A.; and Hakkani-Tür,
  D. 2019.
\newblock Topical-Chat: Towards Knowledge-Grounded Open-Domain Conversations.
\newblock In \emph{Interspeech 2019}.

\bibitem[{Gupta et~al.(2019)Gupta, Kulkarni, Chanda, Rayasam, and
  Lipton}]{2019AmazonQA}
Gupta, M.; Kulkarni, N.; Chanda, R.; Rayasam, A.; and Lipton, Z.~C. 2019.
\newblock AmazonQA: A Review-Based Question Answering Task.

\bibitem[{He et~al.(2021)He, Sun, Yang, Ji, Li, and Xu}]{he2021multi}
He, W.; Sun, Y.; Yang, M.; Ji, F.; Li, C.; and Xu, R. 2021.
\newblock Multi-goal multi-agent learning for task-oriented dialogue with
  bidirectional teacher--student learning.
\newblock \emph{Knowledge-Based Systems}, 213: 106667.

\bibitem[{He et~al.(2020)He, Yang, Yan, Li, Shen, and Xu}]{he2020amalgamating}
He, W.; Yang, M.; Yan, R.; Li, C.; Shen, Y.; and Xu, R. 2020.
\newblock Amalgamating knowledge from two teachers for task-oriented dialogue
  system with adversarial training.
\newblock In \emph{Proceedings of the 2020 Conference on Empirical Methods in
  Natural Language Processing (EMNLP)}, 3498--3507.

\bibitem[{Henderson et~al.(2019)Henderson, Casanueva, Mrk{\v{s}}i{\'c}, Su,
  Wen, and Vuli{\'c}}]{henderson2019convert}
Henderson, M.; Casanueva, I.; Mrk{\v{s}}i{\'c}, N.; Su, P.-H.; Wen, T.-H.; and
  Vuli{\'c}, I. 2019.
\newblock Convert: Efficient and accurate conversational representations from
  transformers.
\newblock \emph{arXiv preprint arXiv:1911.03688}.

\bibitem[{Henderson, Thomson, and Williams(2014)}]{henderson2014second}
Henderson, M.; Thomson, B.; and Williams, J.~D. 2014.
\newblock The second dialog state tracking challenge.
\newblock In \emph{Proceedings of the 15th annual meeting of the special
  interest group on discourse and dialogue (SIGDIAL)}, 263--272.

\bibitem[{Hosseini-Asl et~al.(2020)Hosseini-Asl, McCann, Wu, Yavuz, and
  Socher}]{hosseini2020simple}
Hosseini-Asl, E.; McCann, B.; Wu, C.-S.; Yavuz, S.; and Socher, R. 2020.
\newblock A simple language model for task-oriented dialogue.
\newblock \emph{arXiv preprint arXiv:2005.00796}.

\bibitem[{Jeon and Lee(2021)}]{jeon2021domain}
Jeon, H.; and Lee, G.~G. 2021.
\newblock Domain State Tracking for a Simplified Dialogue System.
\newblock \emph{arXiv preprint arXiv:2103.06648}.

\bibitem[{Jin et~al.(2018)Jin, Lei, Ren, Chen, Liang, Zhao, and
  Yin}]{jin2018explicit}
Jin, X.; Lei, W.; Ren, Z.; Chen, H.; Liang, S.; Zhao, Y.; and Yin, D. 2018.
\newblock Explicit state tracking with semi-supervision for neural dialogue
  generation.
\newblock In \emph{CIKM}, 1403--1412.

\bibitem[{Kingma and Welling(2019)}]{kingma2019introduction}
Kingma, D.~P.; and Welling, M. 2019.
\newblock An introduction to variational autoencoders.
\newblock \emph{arXiv preprint arXiv:1906.02691}.

\bibitem[{Kulh{\'a}nek et~al.(2021)Kulh{\'a}nek, Hude{\v{c}}ek, Nekvinda, and
  Du{\v{s}}ek}]{kulhanek2021augpt}
Kulh{\'a}nek, J.; Hude{\v{c}}ek, V.; Nekvinda, T.; and Du{\v{s}}ek, O. 2021.
\newblock Augpt: Dialogue with pre-trained language models and data
  augmentation.
\newblock \emph{arXiv preprint arXiv:2102.05126}.

\bibitem[{Lee et~al.(2013)}]{lee2013pseudo}
Lee, D.-H.; et~al. 2013.
\newblock Pseudo-label: The simple and efficient semi-supervised learning
  method for deep neural networks.
\newblock In \emph{Workshop on challenges in representation learning, ICML},
  volume~3, 896.

\bibitem[{Lei et~al.(2018)Lei, Jin, Kan, Ren, He, and Yin}]{lei2018sequicity}
Lei, W.; Jin, X.; Kan, M.-Y.; Ren, Z.; He, X.; and Yin, D. 2018.
\newblock Sequicity: Simplifying task-oriented dialogue systems with single
  sequence-to-sequence architectures.
\newblock In \emph{Proceedings of the 56th Annual Meeting of the Association
  for Computational Linguistics (Volume 1: Long Papers)}, 1437--1447.

\bibitem[{Li et~al.(2018)Li, Wang, Sun, Panda, Liu, and Gao}]{li2018microsoft}
Li, X.; Wang, Y.; Sun, S.; Panda, S.; Liu, J.; and Gao, J. 2018.
\newblock Microsoft dialogue challenge: Building end-to-end task-completion
  dialogue systems.
\newblock \emph{arXiv preprint arXiv:1807.11125}.

\bibitem[{Li et~al.(2017)Li, Su, Shen, Li, Cao, and Niu}]{li2017dailydialog}
Li, Y.; Su, H.; Shen, X.; Li, W.; Cao, Z.; and Niu, S. 2017.
\newblock Dailydialog: A manually labelled multi-turn dialogue dataset.
\newblock \emph{arXiv preprint arXiv:1710.03957}.

\bibitem[{Liang et~al.(2021)Liang, Wu, Li, Wang, Meng, Qin, Chen, Zhang, and
  Liu}]{liang2021r}
Liang, X.; Wu, L.; Li, J.; Wang, Y.; Meng, Q.; Qin, T.; Chen, W.; Zhang, M.;
  and Liu, T.-Y. 2021.
\newblock R-Drop: Regularized Dropout for Neural Networks.
\newblock \emph{arXiv preprint arXiv:2106.14448}.

\bibitem[{Lin et~al.(2020)Lin, Madotto, Winata, and Fung}]{lin2020mintl}
Lin, Z.; Madotto, A.; Winata, G.~I.; and Fung, P. 2020.
\newblock Mintl: Minimalist transfer learning for task-oriented dialogue
  systems.
\newblock \emph{arXiv preprint arXiv:2009.12005}.

\bibitem[{Liu et~al.(2018)Liu, Tur, Hakkani-Tur, Shah, and
  Heck}]{liu2018dialogue}
Liu, B.; Tur, G.; Hakkani-Tur, D.; Shah, P.; and Heck, L. 2018.
\newblock Dialogue learning with human teaching and feedback in end-to-end
  trainable task-oriented dialogue systems.
\newblock \emph{arXiv preprint arXiv:1804.06512}.

\bibitem[{Liu et~al.(2021)Liu, Cai, Lin, Ou, Huang, and
  Feng}]{liu2021variational}
Liu, H.; Cai, Y.; Lin, Z.; Ou, Z.; Huang, Y.; and Feng, J. 2021.
\newblock Variational Latent-State GPT for Semi-supervised Task-Oriented Dialog
  Systems.
\newblock \emph{arXiv preprint arXiv:2109.04314}.

\bibitem[{Lubis et~al.(2020)Lubis, Geishauser, Heck, Lin, Moresi, van Niekerk,
  and Gasic}]{lubis-etal-2020-lava}
Lubis, N.; Geishauser, C.; Heck, M.; Lin, H.-c.; Moresi, M.; van Niekerk, C.;
  and Gasic, M. 2020.
\newblock {LAVA}: Latent Action Spaces via Variational Auto-encoding for
  Dialogue Policy Optimization.
\newblock In \emph{Proceedings of the 28th International Conference on
  Computational Linguistics}. Barcelona, Spain (Online): International
  Committee on Computational Linguistics.

\bibitem[{Mehri, Eric, and Hakkani-Tur(2020)}]{mehri2020dialoglue}
Mehri, S.; Eric, M.; and Hakkani-Tur, D. 2020.
\newblock Dialoglue: A natural language understanding benchmark for
  task-oriented dialogue.
\newblock \emph{arXiv preprint arXiv:2009.13570}.

\bibitem[{Mehri et~al.(2019)Mehri, Razumovskaia, Zhao, and
  Eskenazi}]{mehri2019pretraining}
Mehri, S.; Razumovskaia, E.; Zhao, T.; and Eskenazi, M. 2019.
\newblock Pretraining methods for dialog context representation learning.
\newblock \emph{arXiv preprint arXiv:1906.00414}.

\bibitem[{Mehri, Srinivasan, and Eskenazi(2019)}]{mehri2019structured}
Mehri, S.; Srinivasan, T.; and Eskenazi, M. 2019.
\newblock Structured fusion networks for dialog.
\newblock \emph{arXiv:1907.10016}.

\bibitem[{Mosig, Mehri, and Kober(2020)}]{mosig2020star}
Mosig, J.~E.; Mehri, S.; and Kober, T. 2020.
\newblock Star: A schema-guided dialog dataset for transfer learning.
\newblock \emph{arXiv preprint arXiv:2010.11853}.

\bibitem[{Mrki et~al.(2017)Mrki, Séaghdha, Wen, Thomson, and
  Young}]{2017Neural}
Mrki, N.; Séaghdha, D.; Wen, T.~H.; Thomson, B.; and Young, S. 2017.
\newblock Neural Belief Tracker: Data-Driven Dialogue State Tracking.
\newblock In \emph{Proceedings of the 55th Annual Meeting of the Association
  for Computational Linguistics (Volume 1: Long Papers)}.

\bibitem[{Myers, Etchart, and Fulda(2020)}]{myers2020conversational}
Myers, W.; Etchart, T.; and Fulda, N. 2020.
\newblock Conversational Scaffolding: An Analogy-Based Approach to Response
  Prioritization in Open-Domain Dialogs.

\bibitem[{Papineni et~al.(2002)Papineni, Roukos, Ward, and
  Zhu}]{papineni2002bleu}
Papineni, K.; Roukos, S.; Ward, T.; and Zhu, W.-J. 2002.
\newblock BLEU: a method for automatic evaluation of machine translation.
\newblock In \emph{ACL}, 311--318.

\bibitem[{Paul, Goel, and Hakkani-T{\"u}r(2019)}]{paul2019towards}
Paul, S.; Goel, R.; and Hakkani-T{\"u}r, D. 2019.
\newblock Towards universal dialogue act tagging for task-oriented dialogues.
\newblock \emph{arXiv preprint arXiv:1907.03020}.

\bibitem[{Peng et~al.(2020{\natexlab{a}})Peng, Li, Li, Shayandeh, Liden, and
  Gao}]{peng2020soloist}
Peng, B.; Li, C.; Li, J.; Shayandeh, S.; Liden, L.; and Gao, J.
  2020{\natexlab{a}}.
\newblock SOLOIST: Building Task Bots at Scale with Transfer Learning and
  Machine Teaching.
\newblock \emph{arXiv preprint arXiv:2005.05298}.

\bibitem[{Peng et~al.(2020{\natexlab{b}})Peng, Zhu, Li, Li, Li, Zeng, and
  Gao}]{peng2020few}
Peng, B.; Zhu, C.; Li, C.; Li, X.; Li, J.; Zeng, M.; and Gao, J.
  2020{\natexlab{b}}.
\newblock Few-shot natural language generation for task-oriented dialog.
\newblock \emph{arXiv preprint arXiv:2002.12328}.

\bibitem[{Radford et~al.(2019)Radford, Wu, Child, Luan, Amodei, Sutskever
  et~al.}]{gpt2}
Radford, A.; Wu, J.; Child, R.; Luan, D.; Amodei, D.; Sutskever, I.; et~al.
  2019.
\newblock Language models are unsupervised multitask learners.
\newblock \emph{OpenAI blog}, 1: 9.

\bibitem[{Radlinski et~al.(2019)Radlinski, Balog, Byrne, and
  Krishnamoorthi}]{radlinski-etal-2019-coached}
Radlinski, F.; Balog, K.; Byrne, B.; and Krishnamoorthi, K. 2019.
\newblock Coached Conversational Preference Elicitation: A Case Study in
  Understanding Movie Preferences.
\newblock In \emph{Proceedings of the 20th Annual SIGdial Meeting on Discourse
  and Dialogue}, 353--360. Stockholm, Sweden: Association for Computational
  Linguistics.

\bibitem[{Rastogi et~al.(2020)Rastogi, Zang, Sunkara, Gupta, and
  Khaitan}]{rastogi2020towards}
Rastogi, A.; Zang, X.; Sunkara, S.; Gupta, R.; and Khaitan, P. 2020.
\newblock Towards scalable multi-domain conversational agents: The
  schema-guided dialogue dataset.
\newblock In \emph{Proceedings of the AAAI Conference on Artificial
  Intelligence}, volume~34, 8689--8696.

\bibitem[{Roller et~al.(2020)Roller, Dinan, Goyal, Ju, Williamson, Liu, Xu,
  Ott, Shuster, Smith et~al.}]{roller2020recipes}
Roller, S.; Dinan, E.; Goyal, N.; Ju, D.; Williamson, M.; Liu, Y.; Xu, J.; Ott,
  M.; Shuster, K.; Smith, E.~M.; et~al. 2020.
\newblock Recipes for building an open-domain chatbot.
\newblock \emph{arXiv preprint arXiv:2004.13637}.

\bibitem[{Shah et~al.(2018)Shah, Hakkani-Tur, Liu, and
  Tur}]{shah2018bootstrapping}
Shah, P.; Hakkani-Tur, D.; Liu, B.; and Tur, G. 2018.
\newblock Bootstrapping a neural conversational agent with dialogue self-play,
  crowdsourcing and on-line reinforcement learning.
\newblock In \emph{NAACL (Industry Papers)}, 41--51.

\bibitem[{Shalyminov et~al.(2019)Shalyminov, Lee, Eshghi, and Lemon}]{2019Few}
Shalyminov, I.; Lee, S.; Eshghi, A.; and Lemon, O. 2019.
\newblock Few-Shot Dialogue Generation Without Annotated Data: A Transfer
  Learning Approach.
\newblock In \emph{Proceedings of the 20th Annual SIGdial Meeting on Discourse
  and Dialogue}.

\bibitem[{Shu et~al.(2019)Shu, Molino, Namazifar, Xu, Liu, Zheng, and
  Tur}]{shu2019flexibly}
Shu, L.; Molino, P.; Namazifar, M.; Xu, H.; Liu, B.; Zheng, H.; and Tur, G.
  2019.
\newblock Flexibly-structured model for task-oriented dialogues.
\newblock \emph{arXiv preprint arXiv:1908.02402}.

\bibitem[{Su et~al.(2017)Su, Budzianowski, Ultes, Gasic, and
  Young}]{su2017sample}
Su, P.-H.; Budzianowski, P.; Ultes, S.; Gasic, M.; and Young, S. 2017.
\newblock Sample-efficient actor-critic reinforcement learning with supervised
  data for dialogue management.
\newblock \emph{arXiv preprint arXiv:1707.00130}.

\bibitem[{Su et~al.(2021)Su, Shu, Mansimov, Gupta, Cai, Lai, and
  Zhang}]{su2021multitask}
Su, Y.; Shu, L.; Mansimov, E.; Gupta, A.; Cai, D.; Lai, Y.; and Zhang, Y. 2021.
\newblock Multi-Task Pre-Training for Plug-and-Play Task-Oriented Dialogue
  System.
\newblock \emph{CoRR}, abs/2109.14739.

\bibitem[{Sun et~al.(2020)Sun, Wang, Li, Feng, Tian, Wu, and
  Wang}]{sun2020ernie}
Sun, Y.; Wang, S.; Li, Y.; Feng, S.; Tian, H.; Wu, H.; and Wang, H. 2020.
\newblock Ernie 2.0: A continual pre-training framework for language
  understanding.
\newblock In \emph{Proceedings of the AAAI Conference on Artificial
  Intelligence}, volume~34, 8968--8975.

\bibitem[{Tseng et~al.(2021)Tseng, Dai, Kreyssig, and
  Byrne}]{tseng-etal-2021-transferable}
Tseng, B.-H.; Dai, Y.; Kreyssig, F.; and Byrne, B. 2021.
\newblock Transferable Dialogue Systems and User Simulators.
\newblock In \emph{Proceedings of the 59th Annual Meeting of the Association
  for Computational Linguistics and the 11th International Joint Conference on
  Natural Language Processing (Volume 1: Long Papers)}, 152--166. Online:
  Association for Computational Linguistics.

\bibitem[{Verma et~al.(2019)Verma, Kawaguchi, Lamb, Kannala, Bengio, and
  Lopez-Paz}]{verma2019interpolation}
Verma, V.; Kawaguchi, K.; Lamb, A.; Kannala, J.; Bengio, Y.; and Lopez-Paz, D.
  2019.
\newblock Interpolation consistency training for semi-supervised learning.
\newblock \emph{arXiv preprint arXiv:1903.03825}.

\bibitem[{Wang et~al.(2021)Wang, Zhang, Kim, and Gu}]{wang2020modelling}
Wang, J.; Zhang, Y.; Kim, T.-K.; and Gu, Y. 2021.
\newblock Modelling hierarchical structure between dialogue policy and natural
  language generator with option framework for task-oriented dialogue system.
\newblock \emph{ICLR 2021}.

\bibitem[{Wang et~al.(2020)Wang, Tian, Wang, Quan, and
  Yu}]{wang-etal-2020-multi-domain}
Wang, K.; Tian, J.; Wang, R.; Quan, X.; and Yu, J. 2020.
\newblock Multi-Domain Dialogue Acts and Response Co-Generation.
\newblock In \emph{Proceedings of the 58th Annual Meeting of the Association
  for Computational Linguistics}. Online: Association for Computational
  Linguistics.

\bibitem[{Wu et~al.(2020)Wu, Hoi, Socher, and Xiong}]{wu2020tod}
Wu, C.-S.; Hoi, S.; Socher, R.; and Xiong, C. 2020.
\newblock TOD-BERT: pre-trained natural language understanding for
  task-oriented dialogue.
\newblock \emph{EMNLP 2020}.

\bibitem[{Wu and Xiong(2020)}]{wu2020probing}
Wu, C.-S.; and Xiong, C. 2020.
\newblock Probing task-oriented dialogue representation from language models.
\newblock \emph{arXiv preprint arXiv:2010.13912}.

\bibitem[{Xu and Zhao(2021)}]{xu2021dialogue}
Xu, Y.; and Zhao, H. 2021.
\newblock Dialogue-oriented Pre-training.
\newblock \emph{arXiv preprint arXiv:2106.00420}.

\bibitem[{Yang, Li, and Quan(2020)}]{yang2020ubar}
Yang, Y.; Li, Y.; and Quan, X. 2020.
\newblock UBAR: Towards Fully End-to-End Task-Oriented Dialog Systems with
  GPT-2.
\newblock \emph{arXiv preprint arXiv:2012.03539}.

\bibitem[{Yu et~al.(2020)Yu, Zhang, Polozov, Meek, and Awadallah}]{yu2020score}
Yu, T.; Zhang, R.; Polozov, A.; Meek, C.; and Awadallah, A.~H. 2020.
\newblock SCoRe: Pre-Training for Context Representation in Conversational
  Semantic Parsing.
\newblock In \emph{International Conference on Learning Representations}.

\bibitem[{Zeng and Nie(2021)}]{zenginvestigation}
Zeng, Y.; and Nie, J.-Y. 2021.
\newblock An Investigation of Suitability of Pre-Trained Language Models for
  Dialogue Generation--Avoiding Discrepancies.

\bibitem[{Zhang et~al.(2018)Zhang, Dinan, Urbanek, Szlam, Kiela, and
  Weston}]{2018Personalizing}
Zhang, S.; Dinan, E.; Urbanek, J.; Szlam, A.; Kiela, D.; and Weston, J. 2018.
\newblock Personalizing Dialogue Agents: I have a dog, do you have pets too?

\bibitem[{Zhang et~al.(2020{\natexlab{a}})Zhang, Ou, Hu, and
  Feng}]{zhang-etal-2020-probabilistic}
Zhang, Y.; Ou, Z.; Hu, M.; and Feng, J. 2020{\natexlab{a}}.
\newblock A Probabilistic End-To-End Task-Oriented Dialog Model with Latent
  Belief States towards Semi-Supervised Learning.
\newblock In \emph{Proceedings of the 2020 Conference on Empirical Methods in
  Natural Language Processing (EMNLP)}, 9207--9219. Online: Association for
  Computational Linguistics.

\bibitem[{Zhang, Ou, and Yu(2020)}]{zhang2020task}
Zhang, Y.; Ou, Z.; and Yu, Z. 2020.
\newblock Task-oriented dialog systems that consider multiple appropriate
  responses under the same context.
\newblock In \emph{AAAI}, volume~34, 9604--9611.

\bibitem[{Zhang et~al.(2020{\natexlab{b}})Zhang, Sun, Galley, Chen, Brockett,
  Gao, Gao, Liu, and Dolan}]{zhang2019dialogpt}
Zhang, Y.; Sun, S.; Galley, M.; Chen, Y.-C.; Brockett, C.; Gao, X.; Gao, J.;
  Liu, J.; and Dolan, B. 2020{\natexlab{b}}.
\newblock Dialogpt: Large-scale generative pre-training for conversational
  response generation.
\newblock \emph{ACL 2020}.

\end{thebibliography}
\clearpage
\section*{Appendix}
\subsection*{Appendix A}
\label{AppxA}
\subsubsection{A.1. Unified DA Taxonomy.}
\label{AppxA_1}

\begin{figure*}[t]
    \centering
    \includegraphics[width=0.9\textwidth]{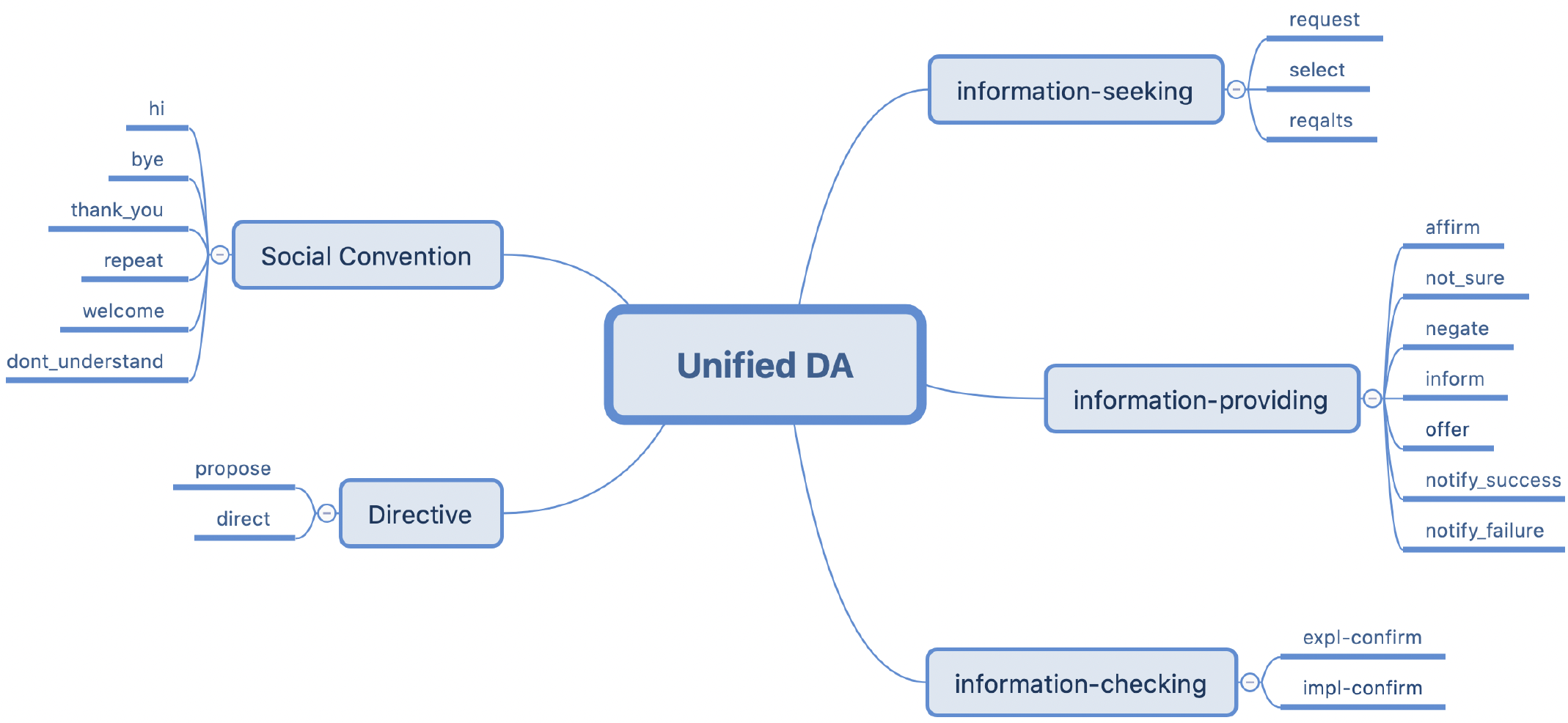}
    \caption{The proposed unified DA taxonomy.}
    \label{fig:DA_schema}
\end{figure*}

The hierarchical structure of our proposed unified DA taxonomy is illustrated in Figure \ref{fig:DA_schema}. There are totally 20 labels.

\paragraph{Social Convention.} This group consists of  DAs about regular actions for social behaviors: \textit{hi}, \textit{bye}, \textit{thank\_you}, \textit{repeat}, \textit{welcome}, \textit{dont\_understand}. 
\begin{itemize}
    \item \textit{hi} means greeting responses, like `hello', `how are you'.
    \item \textit{bye} means the responses for  saying goodbye.
    \item \textit{thank\_you} means the responses for appreciation.
    \item \textit{repeat} means asking the user to repeat what he/she said last turn again.
    \item \textit{welcome} denotes a paragraph of official texts to broadcast the information that the system can offer, like `welcome to Cambridge restaurant, we can help you to order food, you can find restaurants by talking about your favorite foods, area, price range.'
    \item \textit{dont\_understand} means the system can not understand what the user says, which is normal when the user talk about something beyond the semantic scope that the system can process.
\end{itemize}

\paragraph{Directive.} This group consists of  DAs about
 providing suggestions or imperative orders.
 \begin{itemize}
    \item \textit{propose} means suggesting to do/offer/recommend something, in order to make the user consider the performance of a certain action, which the system believes is in the user's interests. For example `How about we find a good place to have fun.'
    \item \textit{direct} means imperative responses that expresses an order, e.g., `you need to open the light before going to bad.'
\end{itemize}

\paragraph{Information Seeking.} This group consists of DAs that perform actions about asking.

\begin{itemize}
    \item \textit{request} means asking the user about specific attributes, like `what area do you like?'
    \item \textit{select} means asking the user to choose a preferred choices from a set of candidates.
     \item \textit{reqalts} means asking the user for more information. e.g., `what else information do you want?'
\end{itemize}

\paragraph{Information Providing.}  This group consists of DAs that provides specific answers to the user.

\begin{itemize}
    \item \textit{affirm} denotes the affirmative responses. e.g., `Yes, it is.'
    \item \textit{not\_sure} means the system is not certain about the user's confirmation. 
     \item \textit{negate} denotes the negating responses. `Noe, it is not.'
     \item \textit{inform} denotes the normal answers to give the information required by the user. e.g.,  `The hotel is in the east area.'
     \item \textit{offer} means the system offer the current searching results from the database that match the user's need.  e.g., `There are 10 restaurants I've found for you.' 
     \item \textit{notify-success} means the system notifies the user that his/her goal is finished successfully . e.g., `Sure, the XXX  is a good one, I've booked it for you.' 
     \item \textit{notify-failure} means the system notifies the user that his/her goal is not finished successfully . e.g., `Sorry, I can not book it for you now, because it is full' 
\end{itemize}

\paragraph{Information Checking.}  This group consists of DAs that the system ask the user about something to confirm whether it is true or correct.

\begin{itemize}
    \item \textit{expl-confirm} means to ask the user explicitly to check something. e.g. `Do you need to cheap restaurant ?'
    \item \textit{impl-confirm} means to check something implicitly, often in a statement that repeats what user says. e.g. `You want a cheap restaurant, OKay.'
\end{itemize}

\subsubsection{A.2. Details for UnDial.}
\label{AppxA_2}
The Detailed statistics are given in Table \ref{tab:UnDialstatictis}.  We totally aggregate 14 dialog corpora from the Internet. 
The processing methods includes: 
(1) Removing the instances
where  there is a URL in utterances.
(2) Removing the instances containing word repetitions of at
least three words
(3) removing non-English sentences.
(4) removing sentence containing special
markers such as “[” or “]”, as this could be markup.
(5) removing offensive language. 
(6) Replacing the non-unicode characters like emojis.



\subsection*{Appendix B}
\label{AppxB}
\subsubsection{B.1. Inputs and Outputs.}
\label{AppxB_1}
Figure \ref{fig:representation} illustrates the input representations in the pre-training stage, we use special tokens [CLS], [BOS] and [EOS] to concatenate sentences in context and the response. Apart from token embeddings, we also have  position emebddings, role embeddings and turn embeddings as in \citet{bao2020plato}.

\begin{table*}[htp]
\centering
\scalebox{1.0}{
\begin{tabular}{lrr}
\hline
Name & \# Dialog & \# Utterance \\ \hline
Reddit \cite{zhang2019dialogpt} & 15,914,021 & 31,908,317 \\
TaskMaster1 \cite{Taskmaster} & 13,215 & 135,176 \\
TaskMaster2 \footnote{https://github.com/google-research-datasets/Taskmaster/tree/master/TM-2-2020} & 17,289  &137,064 \\
TaskMaster3 \footnote{https://github.com/google-research-datasets/Taskmaster/tree/master/TM-3-2020} & 23,789 & 237,617 \\
WOZ \cite{2017Neural}  & 1,200 & 7,624 \\
MetalWOZ \cite{2019Few} & 37,884 & 356,268 \\
ABCD \cite{chen2021action} & 8,034 & 64,500 \\
PersonaChat \cite{2018Personalizing} & 18,876 & 250,634 \\
TopicChat \cite{2019Topical} & 10,784 & 235,434 \\
ChitChat \cite{myers2020conversational} & 7,168 & 258,145 \\
AmazonQA \cite{2019AmazonQA} & 962,260 & 1,924,520 \\
Self-Dialog \cite{fainberg2018talking} & 24,165 & 348,554 \\
Movie-Dialogs \cite{danescu-niculescu-mizil-lee-2011-chameleons} & 220,579 & 441,158 \\
CCPE-M \cite{radlinski-etal-2019-coached} & 502 & 12,000 \\ \hline
Total & 14,021,898 & 35,529,276 \\\hline
\end{tabular}}
\caption{Statistics for each corpus in UnDial.}
\label{tab:UnDialstatictis}
\end{table*}

\begin{figure}[htp]
    \centering
    \includegraphics[width=0.5\textwidth]{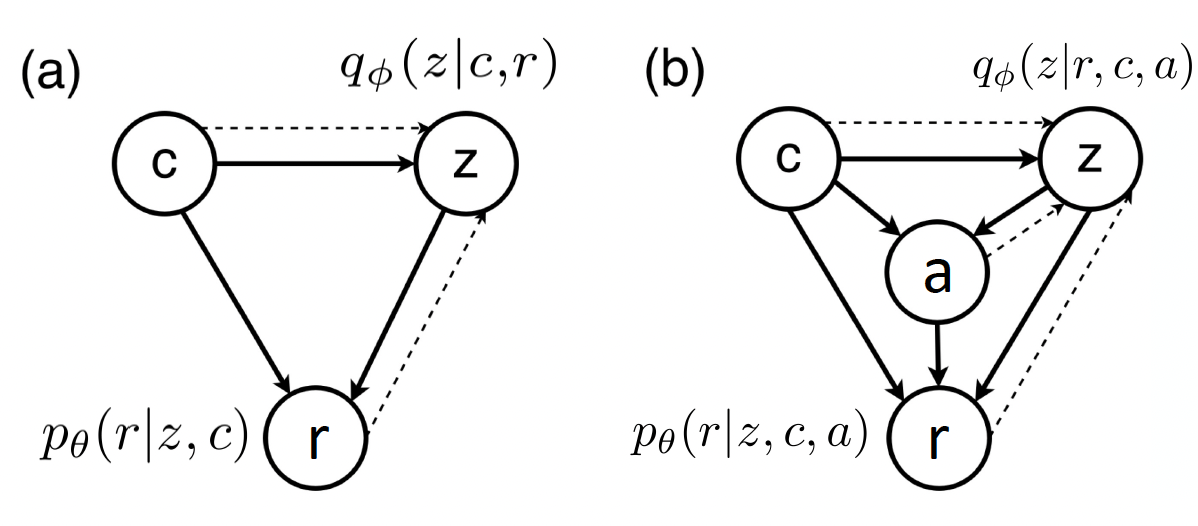}
    \caption{Graphical models of VAE method for semi-supervised pre-training, in which $z$ is the latent variable. The model for unlabeled data is on the left and the model for labeled data is on the right. }
    \label{fig:vae}
\end{figure}

\begin{figure*}[t]
    \centering
    \includegraphics[width=0.7\textwidth]{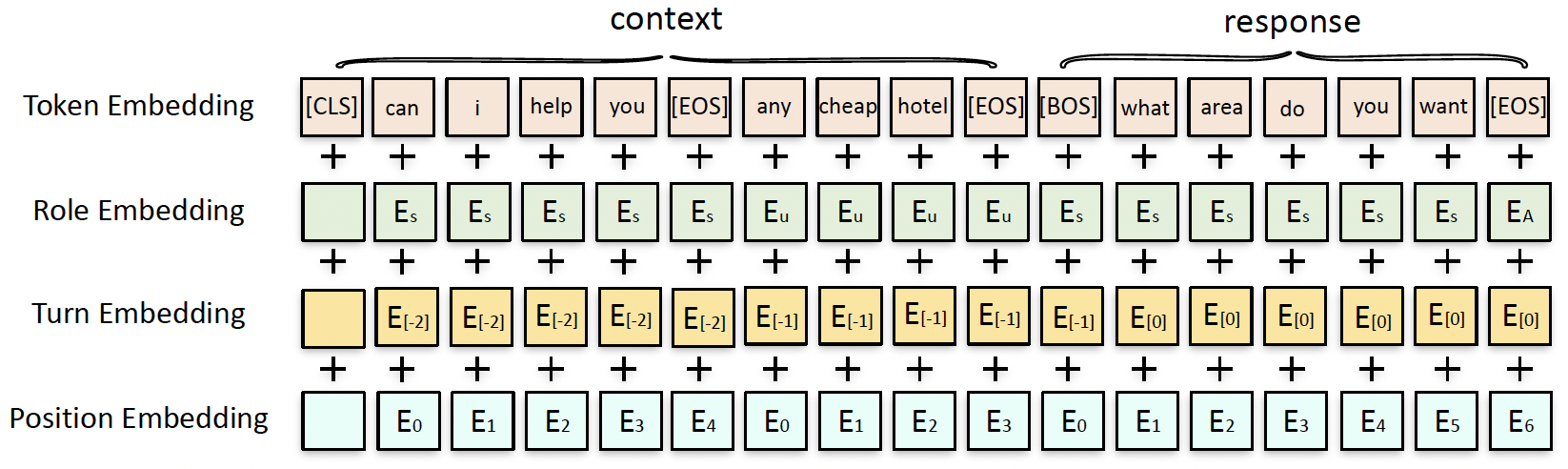}
    \caption{Input representations for the pre-training process.}
    \label{fig:representation}
\end{figure*}

\begin{figure*}[htp]
    \centering
    \includegraphics[width=0.7\textwidth]{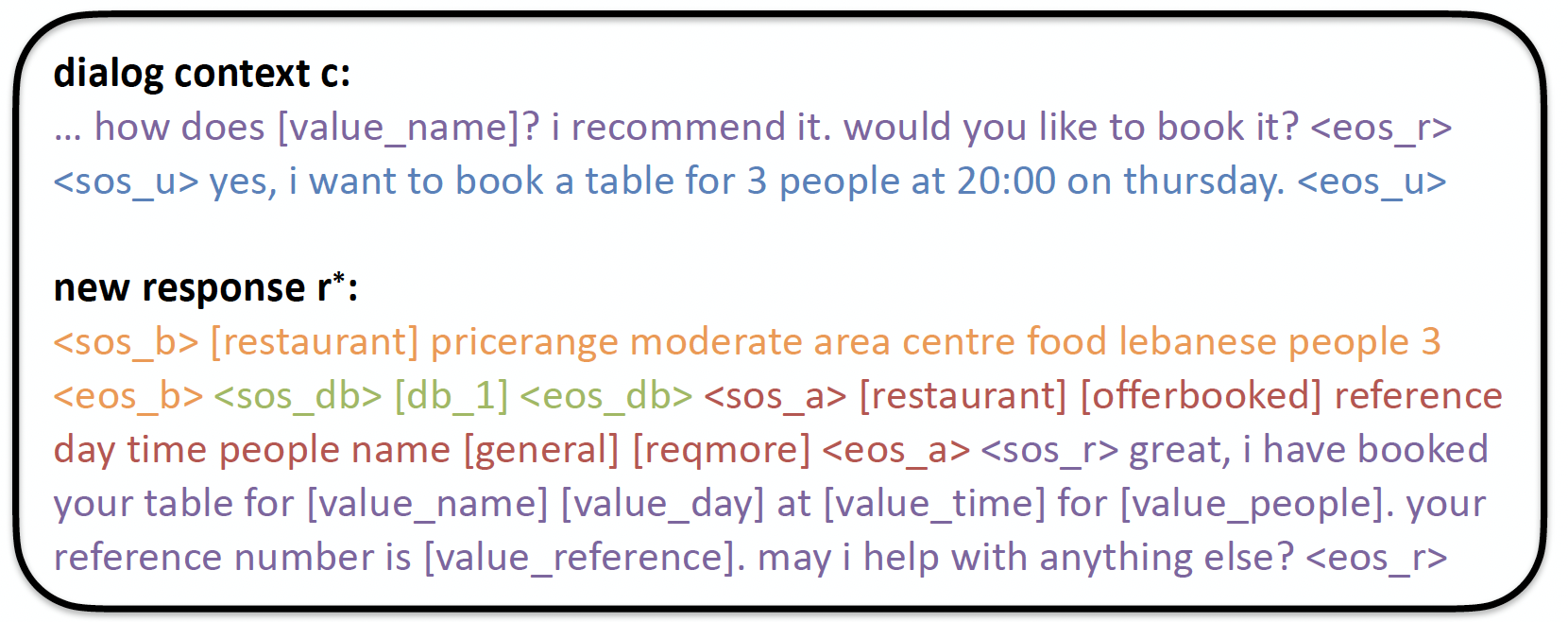}
    \caption{An example of input sequence in downstream tasks. Different colors denote different semantic labels and all labels are converted to text spans: blue for user utterances, orange for belief states, green for database results, red for dialog acts and purple for delexicalized responses.}
    \label{fig:sequence}
\end{figure*}

For the fine-tuning stage, we need to consider the semantic labels, such as `belief states' and `database results', so we add more special tokens to concatenate them as in \citet{yang2020ubar}. Figure \ref{fig:sequence} shows the input sequence of GALAXY in downstream tasks: In-Car and MultiWOZ. 

\subsubsection{B.2. Implementation Details.}
\label{AppxB_2}
We introduce hyper-parameters used in pre-training and fine-tuning as follows.
The number of transformer blocks in GALAXY is 12 and the hidden embedding dimension is 768. The total number of dialog acts $N$ is 20.
In the pre-training stage, GALAXY is initialized with UniLM.
The maximum sequence length of dialog context and response is set to 256 and 50, respectively. The batch size is set to 128 and AdamW optimizer is employed for optimization with an initial learning rate of 1e-5. The dropout rate is set to 0.3 for consistency regularization. 
For semi-supervised pre-training, at each iteration, we mix and shuffle the labeled dataset UniDA and unlabeled dataset UnDial, then randomly sample batches from the mixed corpus as the input of GALAXY. We use a random seed 11, and choose the model checkpoint at the 14th epoch as the final pre-trained model.

For the fine-tuning stage, the maximum sequence length of dialog context and response is set to 1024 and 100 due to longer responses including semantic labels.
The grid search algorithm is applied on the validation set to automatically tune the hyper-parameters.
We use AdamW optimizer with an initial learning rate of 1e-4.
For MultiWOZ dataset, the batch size is set to 32 and the dropout rate is set to 0.1.
For In-Car dataset, the batch size is set to 64 and the dropout rate is set to 0.35.

\subsection*{Appendix C}
\label{AppxC}
\subsubsection{C.1. Other Semi-supervised Pre-training Methods.}
\label{AppxC_1}
\paragraph{Pseudo-Labelling.} 
This method is to train the model with the self-predicted pseudo labels.
Specifically, we first train a model with the same architecture as GALAXY with the loss $\mathcal{L}=\mathcal{L}_\mathrm{DA}+\mathcal{L}_\mathrm{RS}+\mathcal{L}_\mathrm{RG}$ on labeled data, then we use the trained model to predict all pseudo labels on the UnDial. We then train another model the same architecture as GALAXY on all data with the labeled loss in Eq.(\ref{label_loss}). 

\paragraph{Variational Autoencoder (VAE).} 
Figure \ref{fig:vae} shows the framework for the VAE method. We leverage a hidden variable $z$ that has the same size as dialog act $a$. 
For unlabeled data, the generative process of $r$ is (Figure \ref{fig:vae} (a)):
\begin{enumerate}
    \item Sample a latent variable z based on the dialog context $c$ and response $r$ for training: $q_{\phi}(z | c, r)$  while only based on the dialog context $c$ for testing: $p_{\theta}(z | c)$.
    \item Generate the response $r$ based on the dialog context $c$ and latent variable $z$: $p_{\theta}(r | z, c)$.
\end{enumerate}
which is computed as:
\begin{equation}
\begin{aligned}
\mathcal{L}_\mathrm{unlabel} &=
\mathcal{L}(\theta, \phi ; r, c) \\
&=K L\left(q_{\phi}(z | r, c) \| p_{\theta}(z | c)\right) \\
&-\mathbf{E}_{q_{\phi}(z | c, r)}[\log p_{\theta}(r | z, c)]
+\mathcal{L}_\mathrm{RS}
\end{aligned}
\end{equation}

For labeled data, the generative process of $r$ is (Figure \ref{fig:vae} (b)):
\begin{enumerate}
    \item Sample a latent variable z based on the dialog context $c$, response $r$ and dialog act $a$ for training: $q_{\phi}(z | c, r, a)$  while only based on the dialog context $c$ for testing: $p_{\theta}(z | c)$.
    \item Predict the dialog act $a$ based on the dialog context $c$ and latent variable $z$: $p_{\theta} (a | z, c)$.
    \item Generate the response $r$ based on the dialog context $c$, latent variable $z$ and dialog act $a$: $p_{\theta} (r | z, c, a)$.
\end{enumerate}
which is computed as:
\begin{equation}
\begin{aligned}
\mathcal{L}_\mathrm{label} &=
\mathcal{L}(\theta, \phi ; r, c, a) \\
&=K L\left(q_{\phi}(z | r, c, a) \| p_{\theta}(z | c)\right) \\
&-\mathbf{E}_{q_{\phi}(z | c, r, a)}[\log p_{\theta}(r | z, c, a)] \\
&-\mathbf{E}_{q_{\phi}(z | c, r, a)}[\log p_{\theta}(a | z, c)] \\
& +\mathcal{L}_\mathrm{RS}
 + \mathcal{L}_\mathrm{DA} 
\end{aligned}
\end{equation}

To sum up, the final loss $\mathcal{L}_\mathrm{pre}$ for the semi-supervised pre-training is computed as
\begin{equation}
\mathcal{L}_\mathrm{pre}=\mathcal{L}_\mathrm{unlabel} + \mathcal{L}_\mathrm{label}
\end{equation}

\subsection*{Appendix D}
\label{AppxD}
\label{multi_task2}
Table \ref{multiwoz_policy} shows the total end-to-end results given oracle belief states on MultiWOZ2.0 and MultiWOZ2.1.

\begin{table*}[t]
\centering
\scalebox{0.8}{
\begin{tabular}{l|cccc|cccc}
\hline
\multirow{2}{*}{Model} & \multicolumn{4}{c|}{MultiWOZ2.0} & \multicolumn{4}{c}{MultiWOZ2.1} \\ 
 & \texttt{Inform} & \texttt{Success} & \texttt{BLEU} & \texttt{Comb} & \texttt{Inform} & \texttt{Success} & \texttt{BLEU} & \texttt{Comb} \\ \hline
 
SimpleTOD \cite{hosseini2020simple} & 88.9	& 67.1	& 16.9 &	 \multicolumn{1}{c|}{94.9} & 85.1 & 73.5	& 16.22	& 95.52 \\
MarCo \cite{wang-etal-2020-multi-domain} & 92.3&	78.6&	\textbf{20.02}&		 \multicolumn{1}{c|}{105.47} & 92.5	&77.8&	19.54	&104.69	\\
UBAR \cite{yang2020ubar} & 94.0 &	83.6	&17.2&	 \multicolumn{1}{c|}{106} & 92.7 &	81.0	&16.7	&103.55\\
LAVA  \cite{lubis-etal-2020-lava} & \textbf{97.5} &	\textbf{94.8}	&12.1&	 \multicolumn{1}{c|}{108.25} & \textbf{96.39}	&83.57&	14.02&	104\\
HDNO \cite{wang2020modelling} & 96.4 &	84.7	&18.85&	 \multicolumn{1}{c|}{109.4} & 92.8 &	83.0	&18.97	&106.87 \\
JOUST \cite{tseng-etal-2021-transferable} & 94.7	& 86.7	& 18.7	 & \multicolumn{1}{c|}{109.4} & -- & -- & -- & -- \\
GALAXY(w/o pre-train) & 93.6 &	82.6	&18.6 & \multicolumn{1}{c|}{106.7} & 93.7&	83.3&	18.58&	107.08 \\
GALAXY & 94.8 &	85.7&	19.93&	 \multicolumn{1}{c|}{\textbf{110.18}} & 94.8 &	\textbf{86.2}	& \textbf{20.29}	&\textbf{110.79}\\ \hline
\end{tabular}
}
\caption{E2E performances given oracle belief states on MultiWOZ2.0/2.1. All results are from original papers. `w/o pre-train' means using original weights of UniLM for initialization.}
\label{multiwoz_policy}
\end{table*}

\end{document}